\pdfoutput=1
\documentclass[11pt, dvipsnames]{article}

\usepackage[preprint, nohyperref]{acl}

\usepackage{times}
\usepackage{latexsym}

\usepackage[T1]{fontenc}
\usepackage[utf8]{inputenc}

\usepackage{microtype}

\usepackage{inconsolata}

\usepackage{graphicx}

\usepackage{url}
\usepackage[colorlinks,backref=page]{hyperref}
\usepackage{standalone}
\usepackage{amsthm,amsmath,amssymb}
\usepackage{mathtools}
\usepackage{tikz}
\usepackage{booktabs}
\usepackage{caption}
\usepackage{subcaption}
\usepackage[capitalise,noabbrev]{cleveref}
\usepackage{bm}
\usepackage{natbib}
\usepackage{adjustbox}
\usepackage[inline]{enumitem}  % for adjusting lists
\usepackage{colortbl}
\usepackage{multirow}
\usepackage[most]{tcolorbox}
\usepackage{wrapfig}
\usepackage{nicefrac}
\usepackage{soul}

\hypersetup{
  linkcolor=magenta,
  urlcolor=violet,
  citecolor=blue!60!green!90
}

\Crefname{algorithm}{Alg.}{Algs.}
\Crefname{equation}{Eq.}{Eqs.}
\Crefname{figure}{Fig.}{Figs.}
\Crefname{table}{Tbl.}{Tbls.}
\Crefname{section}{\S\!}{\S\!}
\Crefname{subsection}{\S\!}{\S\!}
\Crefname{subsubection}{\S\!}{\S\!}
\Crefname{appendix}{\S\!}{\S\!}
\Crefname{definition}{Def.}{Defs.}
\Crefname{remark}{Rem.}{Rems.}

\usetikzlibrary{shapes.geometric,arrows}

\tikzstyle{attention} = [
  rectangle,
  rounded corners,
  text width=9.25em,
  minimum height=3.5em,
  text centered,
  draw=black,
  fill=blue!25
]

\tikzstyle{dotprod} = [
  rectangle,
  rounded corners,
  text width=6.5em,
  minimum height=2em,
  text centered,
  draw=black,
  fill=red!20
]

\tikzstyle{linear} = [
  rectangle,
  rounded corners,
  minimum width=3.5em,
  minimum height=2em,
  text centered,
  draw=black,
  fill=yellow!30
]

\tikzstyle{linstar} = [
  rectangle,
  rounded corners,
  minimum width=3.5em,
  minimum height=2em,
  text width=3.5em,
  text centered,
  draw=red,
  fill=orange!30
]

\tikzstyle{concat} = [
  rectangle,
  rounded corners,
  minimum width=3.5em, 
  minimum height=2em, 
  text centered,
  draw=black, 
  fill=green!90!blue!20
]

\tikzstyle{slice} = [
  rectangle,
  rounded corners,
  minimum width=3.5em, 
  minimum height=2em, 
  text centered,
  draw=Red, 
  fill=SkyBlue!40
]

\tikzstyle{inpout} = [
  rectangle,
  minimum width=2em, 
  minimum height=2em, 
  text centered, 
]

\tikzstyle{arrow} = [
  thick,
  ->,
  >=stealth
]

\graphicspath{ {figs/} }

\setlist[itemize]{leftmargin=*}
\setlist[enumerate]{leftmargin=*}

\theoremstyle{plain}
\newtheorem{theorem}{Theorem}[section]

\theoremstyle{definition}
\newtheorem{definition}[theorem]{Definition}

\theoremstyle{remark}
\newtheorem{remark}[theorem]{Remark}

\newcommand{\NN}{\mathbb{N}}
\newcommand{\RR}{\mathbb{R}}

\newcommand{\sA}{\mathcal{A}}

\newcommand{\sS}{\mathcal{S}}

\DeclareMathOperator{\softmax}{softmax}

\DeclareMathOperator{\rank}{rank}
\DeclareMathOperator*{\spn}{span}
\newcommand{\tr}[1]{{#1}^{\intercal}}

\newcommand{\lflush}{\hspace{1.2em}}

\definecolor{StnColour}{rgb}{1,0.85,0.87}
\definecolor{OptColour}{rgb}{0.99,1,0.85}
\definecolor{EffColour}{rgb}{0.85,1,0.85}
\definecolor{SupColour}{rgb}{0.85,0.88,1}
\definecolor{GreyColour}{rgb}{0.94,0.94,0.94}

\newcommand{\ours}[1]{\textbf{\underline{#1}}}

\newif\ifcomment
\commenttrue

\ifcomment
	\newcommand\ignacio[1]{\textbf{\textcolor{blue}{IC: #1}}	}
	\newcommand\waleed[1]{\textbf{\textcolor{green}{WI: #1}}	}
	\newcommand\matt[1]{\textbf{\textcolor{blue}{MP: #1}}	}
	
\else	
	\newcommand\ignacio[1]{}
	\newcommand\waleed[1]{}
	\newcommand\matt[1]{}	
		
\fi

\title{Cost-Effective Attention Mechanisms for Low Resource Settings:\\
Necessity \& Sufficiency of 
Linear Transformations}

\author{
  Peyman Hosseini\textsuperscript{1} $\quad$
  Mehran Hosseini\textsuperscript{2} $\quad$
  Ignacio Castro\textsuperscript{1} $\quad$
  Matthew Purver\textsuperscript{1,3} \\
  \textsuperscript{1}School of EECS, Queen Mary University of London, London, UK \\
  \textsuperscript{2}Department of Informatics, King's College London, London, UK \\
  \textsuperscript{3}Department of Knowledge Technologies, Jožef Stefan Institute, Ljubljana, Slovenia \\
  \{s.hosseini, i.castro, m.purver\}@qmul.ac.uk, mehran.hosseini@kcl.ac.uk \\
}

\begin{document}
\maketitle
\begin{abstract}
From natural language processing to vision, Scaled Dot Product Attention (SDPA) is the backbone of most modern deep learning applications.
Unfortunately, its memory and computational requirements can be prohibitive in low-resource settings. 
In this paper, we improve its efficiency without sacrificing its versatility. 
We propose  
three attention variants where we remove consecutive linear transformations or add a novel one, and evaluate them on a range of standard NLP and vision tasks. 
Our proposed models are substantially lighter than standard SDPA (and have 25-50\% fewer parameters). 
We show that the performance cost of these changes is negligible relative to size reduction and that in one case (Super Attention) we succeed in outperforming SDPA by up to 10\% while improving its speed and reducing its parameters by 25\%.
%
% We also introduce Super Attention, which outperforms 
% standard SPDA in both modalities by up to a 10\% and reduces and is xx times faster than SPDA thanks to having one matrix multiplication per head and 25\% parameters less.

%   while having one fewer matrix multiplication per head and 25\% fewer
%   parameters than standard SPDA

% Super Attention introduces a new linear transformation
%   on the values, transforming them from the left. It outperforms
%   standard SPDA in both modalities by up to 10\%
%   while having one fewer matrix multiplication per head and 25\% fewer
%   parameters than standard SPDA. Consequently, it is also faster than standard SDPA, making it particularly suitable for low-resource deployments.
\end{abstract}

\section{Introduction}
\label{sec: Introduction}
Few ideas have had as profound an effect on the field of
\emph{Artificial Intelligence} (\emph{AI}) as the \emph{attention
  mechanism} \citep{BahdanauCB14}. Introduced as a method to improve
machine translation, the attention mechanism revolutionized the way
neural networks process and interpret data. It mimics a form of cognitive attention in humans by allowing models to
focus on specific parts of the input while disregarding irrelevant
information. This enhanced the capability and efficiency of Language Models
(LM)~and~paved the way for the development of advanced AI
architectures like the Transformer model \citep{Vaswani+17}.

\begin{figure*}[th]
  \vspace{-0.2em}
  \centering 
  \begin{subfigure}[t]{0.23\textwidth}
    \centering
    \begin{tikzpicture}[node distance=5em]
% Main Attention Node
\node (start3) [attention, xshift=0.6em, yshift=0.6em, opacity=0.3] {};
\node (start2) [attention, xshift=0.3em, yshift=0.3em, opacity=0.6] {};
\node (start1) [attention] {Softmax of \\Scaled Dot-Product};

% ********** Below Main ********** %
% ******************************** %
% K Processing
\node (fn-k3) [linear, below of=start1, xshift=3.3em, yshift=0.6em, fill opacity=0.3] {};
\node (fn-k2) [linear, below of=start1, xshift=3.0em, yshift=0.3em, fill opacity=0.6] {};
\node (fn-k1) [linear, below of=start1, xshift=2.7em] {Linear};
% V processing
\node (fn-v3) [linear, right of=fn-k1, xshift=0.6em, yshift=0.6em, fill opacity=0.3] {};
\node (fn-v2) [linear, right of=fn-k1, xshift=0.3em, yshift=0.3em, fill opacity=0.6] {};
\node (fn-v1) [linear, right of=fn-k1] {Linear};
% Q processing
\node (fn-q3) [linear, left of=fn-k1, xshift=0.6em, yshift=0.6em, fill opacity=0.3] {};
\node (fn-q2) [linear, left of=fn-k1, xshift=0.3em, yshift=0.3em, fill opacity=0.6] {};
\node (fn-q1) [linear, left of=fn-k1] {Linear};
% ******************************** %
% ******************************** %

% ********** Above Main ********** %
% ******************************** %
\node (a-dot-v3) [dotprod, above of=start3, opacity=0.3] {};
\node (a-dot-v2) [dotprod, above of=start2, opacity=0.6] {};
\node (a-dot-v1) [dotprod, above of=start1] {Dot Product};
\node (concat) [concat, above of=a-dot-v1, yshift=-0.4em] {Concat};
\node (fn-o) [linear, above of=concat, yshift=-1em] {Linear};
\node (out) [inpout, above of=fn-o, yshift=-1em] {$O$};
% ******************************** %
% ******************************** %

% ****** Arrows Below Main ******* %
% ******************************** %
% QKV
\node (k) [inpout, below of=fn-k1, yshift=1.0em] {$K$};
\node (v) [inpout, below of=fn-v1, yshift=1.0em] {$V$};
\node (q) [inpout, below of=fn-q1, yshift=1.0em] {$Q$};

% QKV processing -> Attention
% K -> Attention
\draw [arrow] (fn-k1) -- (start1.south -| fn-k1.north);
\draw [thick, opacity=0.6] (fn-k2) -- (start1.south -| fn-k2.north);
\draw [thick, opacity=0.3] (fn-k3) -- (start1.south -| fn-k3.north);
% Q -> Attention
\draw [arrow] (fn-q1.north) -- (start1.south -| fn-q1.north);
\draw [thick, opacity=0.6] (fn-q2) -- (start1.south -| fn-q2.north);
\draw [thick, opacity=0.3] (fn-q3) -- (start1.south -| fn-q3.north);
% V -> Dot Product
\draw [arrow, rounded corners=0.5em] (fn-v1.north) |- (a-dot-v1.east);
\draw [arrow, rounded corners=0.5em, opacity=0.6] (fn-v2.north) |- (a-dot-v2.east);
\draw [arrow, rounded corners=0.5em, opacity=0.3] (fn-v3.north) |- (a-dot-v3.east);
% Inpus
\draw [arrow] (k) -- (fn-k1);
\draw [arrow] (v) -- (fn-v1);
\draw [arrow] (q) -- (fn-q1);
% ******************************** %
% ******************************** %

% ****** Arrows Above Main ******* %
% ******************************** %

% A -> A.V
\draw [arrow]  (start1.north) -- (a-dot-v1);
\draw [thick, opacity=0.6] (start2.north) -- (a-dot-v1.south -| a-dot-v2.south);
\draw [thick, opacity=0.3] (start3.north) -- (a-dot-v1.south -| a-dot-v3.south);

% A.V -> Concat
\draw [arrow] ([xshift=-0.4em]a-dot-v1.north) -- ([xshift=-0.4em]concat.south -| a-dot-v1.north);
\draw [arrow, opacity=0.6] ([xshift=-0.2em]a-dot-v2.north) -- ([xshift=-0.2em]concat.south -| a-dot-v2.north);
\draw [arrow, opacity=0.3] ([xshift=0.0em]a-dot-v3.north) -- ([xshift=0.0em]concat.south -| a-dot-v3.north);

% Concat -> Output Linear
\draw [arrow] (concat) -- (fn-o);

% Output
\draw [arrow] (fn-o) -- (out);
% ******************************** %
% ******************************** %
\end{tikzpicture}
    % \vspace{-0.2em}
    \subcaption{Stn. Attention}
    \label{subfig: Standard Attention}
  \end{subfigure}
  \hfill
  \begin{subfigure}[t]{0.23\textwidth}
    \centering
    \begin{tikzpicture}[node distance=5em]
% Main Attention Node
\node (start3) [attention, xshift=0.6em, yshift=0.6em, opacity=0.3] {};
\node (start2) [attention, xshift=0.3em, yshift=0.3em, opacity=0.6] {};
\node (start1) [attention] {Softmax of \\Scaled Dot-Product};

% ********** Below Main ********** %
% ******************************** %
% K Processing
\node (fn-k3) [linear, below of=start1, xshift=3.3em, yshift=0.6em, fill opacity=0.3] {};
\node (fn-k2) [linear, below of=start1, xshift=3.0em, yshift=0.3em, fill opacity=0.6] {};
\node (fn-k1) [linear, below of=start1, xshift=2.7em] {Linear};
% V processing
\node (fn-v3) [slice, right of=fn-k1, xshift=0.6em, yshift=0.6em, fill opacity=0.3] {};
\node (fn-v2) [slice, right of=fn-k1, xshift=0.3em, yshift=0.3em, fill opacity=0.6] {};
\node (fn-v1) [slice, right of=fn-k1] {\ours{Slice}};
% Q processing
\node (fn-q3) [linear, left of=fn-k1, xshift=0.6em, yshift=0.6em, fill opacity=0.3] {};
\node (fn-q2) [linear, left of=fn-k1, xshift=0.3em, yshift=0.3em, fill opacity=0.6] {};
\node (fn-q1) [linear, left of=fn-k1] {Linear};
% ******************************** %
% ******************************** %

% ********** Above Main ********** %
% ******************************** %
\node (a-dot-v3) [dotprod, above of=start3, opacity=0.3] {};
\node (a-dot-v2) [dotprod, above of=start2, opacity=0.6] {};
\node (a-dot-v1) [dotprod, above of=start1] {Dot Product};
\node (concat) [concat, above of=a-dot-v1, yshift=-0.4em] {Concat};
\node (fn-o) [linear, above of=concat, yshift=-1em] {Linear};
\node (out) [inpout, above of=fn-o, yshift=-1em] {$O$};
% ******************************** %
% ******************************** %

% ****** Arrows Below Main ******* %
% ******************************** %
% QKV
\node (k) [inpout, below of=fn-k1, yshift=1.0em] {$K$};
\node (v) [inpout, below of=fn-v1, yshift=1.0em] {$V$};
\node (q) [inpout, below of=fn-q1, yshift=1.0em] {$Q$};

% QKV processing -> Attention
% K -> Attention
\draw [arrow] (fn-k1) -- (start1.south -| fn-k1.north);
\draw [thick, opacity=0.6] (fn-k2) -- (start1.south -| fn-k2.north);
\draw [thick, opacity=0.3] (fn-k3) -- (start1.south -| fn-k3.north);
% Q -> Attention
\draw [arrow] (fn-q1.north) -- (start1.south -| fn-q1.north);
\draw [thick, opacity=0.6] (fn-q2) -- (start1.south -| fn-q2.north);
\draw [thick, opacity=0.3] (fn-q3) -- (start1.south -| fn-q3.north);
% V -> Dot Product
\draw [arrow, rounded corners=0.5em] (fn-v1.north) |- (a-dot-v1.east);
\draw [arrow, rounded corners=0.5em, opacity=0.6] (fn-v2.north) |- (a-dot-v2.east);
\draw [arrow, rounded corners=0.5em, opacity=0.3] (fn-v3.north) |- (a-dot-v3.east);
% Inpus
\draw [arrow] (k) -- (fn-k1);
\draw [arrow] (v) -- (fn-v1);
\draw [arrow] (q) -- (fn-q1);
% ******************************** %
% ******************************** %

% ****** Arrows Above Main ******* %
% ******************************** %

% A -> A.V
\draw [arrow]  (start1.north) -- (a-dot-v1);
\draw [thick, opacity=0.6] (start2.north) -- (a-dot-v1.south -| a-dot-v2.south);
\draw [thick, opacity=0.3] (start3.north) -- (a-dot-v1.south -| a-dot-v3.south);

% A.V -> Concat
\draw [arrow] ([xshift=-0.4em]a-dot-v1.north) -- ([xshift=-0.4em]concat.south -| a-dot-v1.north);
\draw [arrow, opacity=0.6] ([xshift=-0.2em]a-dot-v2.north) -- ([xshift=-0.2em]concat.south -| a-dot-v2.north);
\draw [arrow, opacity=0.3] ([xshift=0.0em]a-dot-v3.north) -- ([xshift=0.0em]concat.south -| a-dot-v3.north);

% Concat -> Output Linear
\draw [arrow] (concat) -- (fn-o);

% Output
\draw [arrow] (fn-o) -- (out);
% ******************************** %
% ******************************** %
\end{tikzpicture}
    % \vspace{-0.2em}
    \subcaption{Opt. Attention}
    \label{subfig: Optimised Attention}
  \end{subfigure}
  \hfill
  \begin{subfigure}[t]{0.23\textwidth}
    \centering
    \begin{tikzpicture}[node distance=5em]
% Main Attention Node
\node (start3) [attention, xshift=0.6em, yshift=0.6em, opacity=0.3] {};
\node (start2) [attention, xshift=0.3em, yshift=0.3em, opacity=0.6] {};
\node (start1) [attention] {Softmax of \\Scaled Dot-Product};

% ********** Below Main ********** %
% ******************************** %
% K Processing
\node (fn-k3) [slice, below of=start1, xshift=3.3em, yshift=0.6em, fill opacity=0.3] {};
\node (fn-k2) [slice, below of=start1, xshift=3.0em, yshift=0.3em, fill opacity=0.6] {};
\node (fn-k1) [slice, below of=start1, xshift=2.7em] {\ours{Slice}};
% V processing
\node (fn-v3) [slice, right of=fn-k1, xshift=0.6em, yshift=0.6em, fill opacity=0.3] {};
\node (fn-v2) [slice, right of=fn-k1, xshift=0.3em, yshift=0.3em, fill opacity=0.6] {};
\node (fn-v1) [slice, right of=fn-k1] {\ours{Slice}};
% Q processing
\node (fn-q3) [linear, left of=fn-k1, xshift=0.6em, yshift=0.6em, fill opacity=0.3] {};
\node (fn-q2) [linear, left of=fn-k1, xshift=0.3em, yshift=0.3em, fill opacity=0.6] {};
\node (fn-q1) [linear, left of=fn-k1] {Linear};
% ******************************** %
% ******************************** %

% ********** Above Main ********** %
% ******************************** %
\node (a-dot-v3) [dotprod, above of=start3, opacity=0.3] {};
\node (a-dot-v2) [dotprod, above of=start2, opacity=0.6] {};
\node (a-dot-v1) [dotprod, above of=start1] {Dot Product};
\node (concat) [concat, above of=a-dot-v1, yshift=-0.4em] {Concat};
\node (fn-o) [linear, above of=concat, yshift=-1em] {Linear};
\node (out) [inpout, above of=fn-o, yshift=-1em] {$O$};
% ******************************** %
% ******************************** %

% ****** Arrows Below Main ******* %
% ******************************** %
% QKV
\node (k) [inpout, below of=fn-k1, yshift=1.0em] {$K$};
\node (v) [inpout, below of=fn-v1, yshift=1.0em] {$V$};
\node (q) [inpout, below of=fn-q1, yshift=1.0em] {$Q$};

% QKV processing -> Attention
% K -> Attention
\draw [arrow] (fn-k1) -- (start1.south -| fn-k1.north);
\draw [thick, opacity=0.6] (fn-k2) -- (start1.south -| fn-k2.north);
\draw [thick, opacity=0.3] (fn-k3) -- (start1.south -| fn-k3.north);
% Q -> Attention
\draw [arrow] (fn-q1.north) -- (start1.south -| fn-q1.north);
\draw [thick, opacity=0.6] (fn-q2) -- (start1.south -| fn-q2.north);
\draw [thick, opacity=0.3] (fn-q3) -- (start1.south -| fn-q3.north);
% V -> Dot Product
\draw [arrow, rounded corners=0.5em] (fn-v1.north) |- (a-dot-v1.east);
\draw [arrow, rounded corners=0.5em, opacity=0.6] (fn-v2.north) |- (a-dot-v2.east);
\draw [arrow, rounded corners=0.5em, opacity=0.3] (fn-v3.north) |- (a-dot-v3.east);
% Inpus
\draw [arrow] (k) -- (fn-k1);
\draw [arrow] (v) -- (fn-v1);
\draw [arrow] (q) -- (fn-q1);
% ******************************** %
% ******************************** %

% ****** Arrows Above Main ******* %
% ******************************** %

% A -> A.V
\draw [arrow]  (start1.north) -- (a-dot-v1);
\draw [thick, opacity=0.6] (start2.north) -- (a-dot-v1.south -| a-dot-v2.south);
\draw [thick, opacity=0.3] (start3.north) -- (a-dot-v1.south -| a-dot-v3.south);

% A.V -> Concat
\draw [arrow] ([xshift=-0.4em]a-dot-v1.north) -- ([xshift=-0.4em]concat.south -| a-dot-v1.north);
\draw [arrow, opacity=0.6] ([xshift=-0.2em]a-dot-v2.north) -- ([xshift=-0.2em]concat.south -| a-dot-v2.north);
\draw [arrow, opacity=0.3] ([xshift=0.0em]a-dot-v3.north) -- ([xshift=0.0em]concat.south -| a-dot-v3.north);

% Concat -> Output Linear
\draw [arrow] (concat) -- (fn-o);

% Output
\draw [arrow] (fn-o) -- (out);
% ******************************** %
% ******************************** %
\end{tikzpicture}
    % \vspace{-0.2em}
    \subcaption{Eff. Attention}
    \label{subfig: Efficient Attention}
  \end{subfigure}
  \hfill
  \begin{subfigure}[t]{0.23\textwidth}
    \centering
    \includestandalone[width=\textwidth]{archs/sup_att}
    % \vspace{-0.2em}
    \subcaption{Sup. Attention}
    \label{subfig: Super Attention}
  \end{subfigure}
  \vspace{-0.4em}
  \caption{Standard multi-head scaled dot product attention
    (\ref{subfig: Standard Attention}) alongside the proposed
    variations: Optimized Attention (\ref{subfig: Optimised
      Attention}), Efficient Attention (\ref{subfig: Efficient
      Attention}), and Super Attention (\ref{subfig: Super
      Attention}). The ``Linear'' block denotes a linear
    transformation right while ``Linear*'' denotes a linear
    transformation from left.}
  \label{fig: Attention Flowcharts}
  \vspace{-0.9em}
\end{figure*}
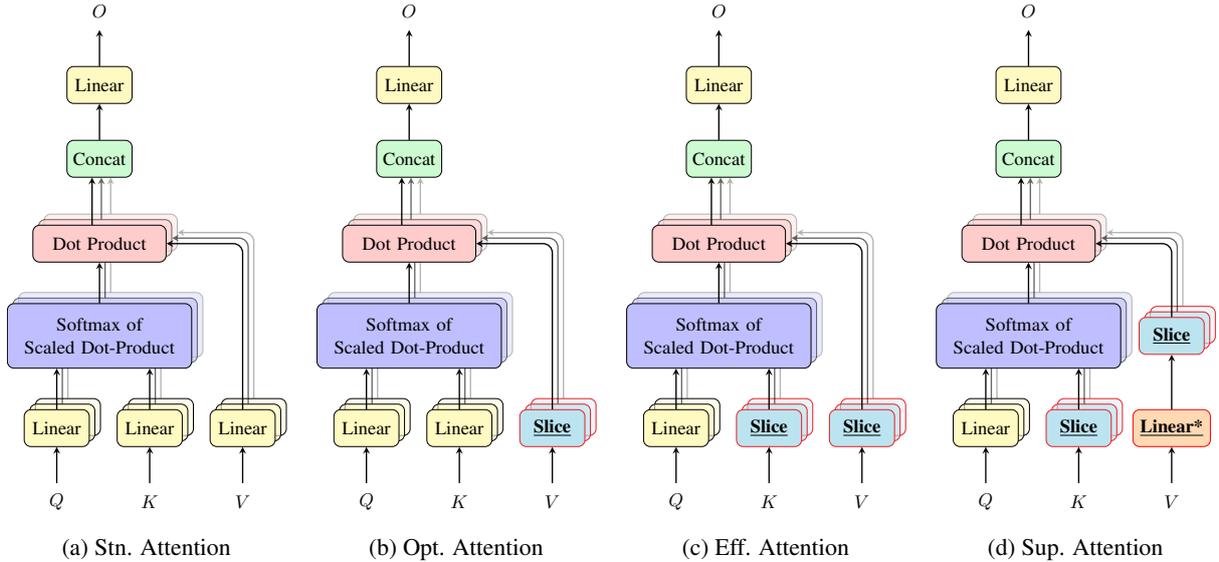

These advances have had far-reaching impacts, extending beyond Natural
Language Processing (NLP) to  areas such as image recognition
\citep{Dosovitskiy+21}, autonomous systems \citep{Mott+19}, healthcare
\citep{Choi+16}, and multi-modal application \cite{XuZC23}.

The formulation of SDPA in all these domains has undergone very little
change compared to the original formulation of
\citet{Vaswani+17}. Instead, the
prevailing maxim has been ``the bigger the better", and  
Large Language Models
(LLM), such as Llama 3 \citep{Touvron+23llama, Touvron+23llama2},
GPT-4 \citep{GPT4}, and Gemini \citep{Gemini23} have demonstrated
unprecedented capabilities in multi-modal domains.

The behemothic sizes of these models have introduced numerous
challenges. Expensive and slow training and inference have resulted in high carbon emissions
\citep{Dhar20}; and such models are impossible not only to
run but even to store on edge devices such as smartphones, consumer
laptops, and even powerful personal workstations.

Numerous attempts have been made to address this
problem using post-training techniques, like quantization
\citep{Jacob+18}, Low-Rank Adaptation (LoRA) \citep{Hu+22}, Quantized
LoRA (QLoRA) \citep{DettmersPHZ23}, and sparsification
\citep{Ashkboos+24}.  
Others have attempted to optimise the
speed and GPU utilization of attention-based models, e.g.,  Flash Attention 1--3 \citep{Dao+22,Dao23,Shah+24}.
However, all these approaches strive to improve the performance of
attention-based models but without altering the attention mechanism. 

In
this paper, we propose a different approach: modifying the attention mechanism itself.
We employ two intuitive principles to design our alternative attention mechanism:
\begin{enumerate*}
\item[\textbf{(1)}]\label{Princ: Linear Combo} two consecutive linear transformations
  do not introduce non-linearity, and
\item[\textbf{(2)}] a learnable linear kernel between each
  two inputs of SDPA enhances learning.
  \label{Princ: Kernel In Between}
\end{enumerate*}
We leverage these two principles to  propose 3 SDPA variants:
\vspace{-0.4em}
\begin{itemize}
\item[$\diamond$] \textbf{Optimized Attention} 
(\cref{subsec:
    Merging WV and WO}, \cref{subfig: Optimised Attention}),
 replaces \(W^V\) linear transformation with a  slicing operation (Principle~1),  reducing the %number of 
 parameters in the attention layer by
  25\% and its computational cost by \(h\) matrix multiplications, where
  \(h\) is the number of heads.  Optimized Attention reduces the
  inference time by 2.5--7.5\%, with little or no performance degradation (\cref{sec: Evaluation}).
\vspace{-0.3em}
\item[$\diamond$] \textbf{Efficient Attention} (\cref{subsec:
    Merging WQ and WK}, \cref{subfig: Efficient Attention})
    replaces \(W^V\) and \(W^K\) linear transformations by  slicing operations (Principle~1). This reduces the %number of 
    parameters in the attention layer by 50\% and its computational cost by \(2h\) matrix multiplications, where \(h\) is the number of heads. Efficient Attention reduces the
  inference time by 5--15\%, with no/little performance degradation (\cref{sec: Evaluation}).
\vspace{-0.3em}
\item[$\diamond$] \textbf{Super Attention} (\cref{subsec:
    Introducing WA}, \cref{subfig: Super Attention}) introduces a new linear operation \(W^A\) (Principle~2), which transforms the values \(V\) from the left. 
    Super Attention can be used on top of standard or Optimized attentions (i.e., without replacing \(W^V\) and \(W^K\)). For simplicity, we build Super Attention on top of Efficient Attention. 
    Super Attention reduces the attention layer's size by \(\sim25\%\) (depending on the attention's context length) and its computational cost by \(h\) matrix multiplications.  Super Attention outperforms standard attention by 2--10\% in NLP and vision tasks and reduces the training and inference time by 2.5--10\% (\cref{sec: Evaluation}).
\end{itemize}
\vspace{-0.4em}

Our evaluation is comprehensive and compares our proposed attention models with SDPA
in the \emph{self-attention} setting in transformers
for multiple datasets and
for 4 different tasks, including:  \textbf{(1)} \emph{Natural
Language Sentiment Classification} on IMDB and Amazon Reviews datasets;
\textbf{(2)} \emph{Machine Translation} (\emph{NMT}) on the combined Europarl and Anki
English-to-Spanish translation dataset; \textbf{(3)} \emph{Generative Language Modeling and Natural Language Inference (NLI)} using NanoGPT \cite{Karpathy22} on the OpenWebText dataset; and to show how these architectural changes generalize to transformers for other modalities, we do complementary experiments for \textbf{(1)} \emph{image
classification} on MNIST, CIFAR100, and ImageNet datasets;

% We evaluate SDPA and our proposed variations %\textcolor{red}{
% in the \emph{self-attention} setting in transformers
% %}
% on \textbf{(1)} \emph{image
% classification} on MNIST, CIFAR100, and ImageNet datasets, \textbf{(2)} \emph{natural
% language sentiment classification} on IMDB and Amazon Reviews datasets, \textbf{(3)} \emph{Neural Machine Translation} (\emph{NMT}) on the combined Europarl and Anki
% English-to-Spanish translation dataset, and \textbf{(4)} \emph{generative language modelling} using Andrea Karpathy's NanoGPT on the OpenWebText dataset.

%%% Local Variables:
%%% mode: latex
%%% TeX-master: "../main"
%%% End:

%
%%
\section{Preliminaries}
\label{sec: Preliminaries}
We start by introducing the notation we use
throughout the paper. For natural numbers \(d_m, d_k \in \NN\), we
denote the \(d_m\)-dimensional real \emph{vectors space} by
\(\RR^{d_m}\) and the set of all real \(d_m \times d_k\)
\emph{matrices} by \(\RR^{d_m \times d_k}\), noting  that all matrices
can be regarded as 2D \emph{tensors} and vice versa. Given a set
\(\sA \subseteq \RR^{d_m}\), we denote the smallest real vector space
containing \(\sA\) by \(\spn(\sA)\). Similarly, given a matrix
\(W \in \RR^{d_m \times d_k}\), we denote the smallest real vector
space containing the columns of \(W\)'s by \(\spn(W)\). For a
\emph{subspace} \(\sS \leq \RR^{d_m}\), the \emph{dimension} of
\(\sS\), denoted \(\dim(\sS)\), is the size of the largest
\emph{linearly independent} set in \(\sS\). The \emph{rank} of a
matrix \(W \in \RR^{d_m \times d_k}\), denoted \(\rank(W)\), is the
number of linearly independent columns (or rows) in \(W\). The
rank-nullity theorem implies that \(\rank(W) = \dim(\spn(W))\) and
\(\rank(W) \leq \min(d_m, d_k)\).\footnote{For details see \citep[Chapters 2 \& 4]{Meyer23}.}

We use the widely-adopted definition of SDPA %the attention mechanism 
as implemented in SotA open-source models such as Llama-3 and Mistral, and
machine learning frameworks like Torch and JAX. For consistency, we use the same notation as \citep{Vaswani+17}.
\begin{definition}[Standard Attention]
  \label{def: Legacy Attention}
  The (\emph{multi-head}) \emph{scaled dot-product attention} on
  \emph{input} tensors \(Q, K, V \in \RR^{\ell \times d_m}\) is
  defined as: %\medskip
  \vspace{-0.2em}
  \begin{center}
  \begin{tcolorbox}[colback=GreyColour,colframe=black,arc=5pt,boxrule=1pt,width=19em, top=-13pt, left=0pt, bottom=-1pt]
  \begin{align}
    O\  & = (H_1 \ H_2 \ \cdots \ H_h) W^O, \label{eq: Legacy O}\\
    H_i & = S_i V'_i, \label{eq: Legacy H}\\
    S_i & = \softmax(\frac{Q'_i\tr{K'}_i}{\sqrt{d_k}}), \label{eq: Legacy S}\\
    V'_i & = V W^V_i, \label{eq: Legacy V}\\
    K'_i & = K W^K_i, \label{eq: Legacy K}\\
    Q'_i & = Q W^Q_i, \label{eq: Legacy Q}
  \end{align}
  \end{tcolorbox}
  \end{center}

  where \(O\) is the \emph{output}; \(Q'_i, K'_i, V'_i, S_i\), and
  \(H_i\) are the \emph{query}, \emph{key}, \emph{value},
  \emph{attention score}, and \emph{head value} of the \(i\)-th
  \emph{head}, respectively. The natural numbers \(\ell, d_m\) and
  \(h\) are the \emph{context length}, \emph{model dimension}, and
  \emph{number of heads}, respectively.  Moreover,
  \(W^Q_i, W^K_i \in \RR^{d_m \times d_k}\) and
  \(W^V_i \in \RR^{d_m \times d_v}\), where \(d_k\) and \(d_v\) are
  the \emph{key} and \emph{value dimensions}, respectively.

  Parameters \(d_m, d_k, d_v\) and \(h\) are often chosen so that
  \(d_k = d_v = d_m / h\), and in recent models, including
  SotA Transformer models, \(Q, K\), and \(V\) are set to \(X\), a single
  input tensor; whereby, the attention mechanism is called
  \emph{self-attention}.
\end{definition}
%

%%% Local Variables:
%%% mode: latex
%%% TeX-master: "../main"
%%% End:

%
%%
\section{Revising the Attention Mechanism}
\label{sec: Theory}
We introduce our three proposed attention variants and discuss the motivation behind each of them.

% We now discuss our motivation for revisiting the attention mechanism. It is important to note
% these variants are not mathematically equivalent to standard attention,
% and our goal here is to justify the choices of variants discussed in
% this paper. These variants are \emph{Optimized Attention},
% \emph{Efficient Attention}, and \emph{Super Attention}, which we introduce in
% Sections~\ref{subsec: Merging WV and WO}, \ref{subsec: Merging WQ and WK}, and \ref{subsec: Introducing WA}, respectively.

% of the attention mechanism and present enhancements to the attention
% mechanism to make it \emph{more efficient} (in terms of \emph{number
%   of parameters} and \emph{computation cost}) and \emph{more capable}
% (in terms of attaining \emph{higher accuracies} and \emph{lower
%   losses}). In particular, we introduce \emph{Optimized Attention} in
% \cref{subsec: Merging WV and WO}, \emph{Efficient Attention} in
% \cref{subsec: Merging WQ and WK}, and \emph{Super Attention} in
% \cref{subsec: Introducing WA}. We provide the mathematical motivation
% behind each of the mechanisms in the corresponding sections and
% evaluate all of them in \cref{sec: Evaluation}.

%
%%
\subsection{Optimized Attention: Absorbing \(W^V_i\)'s into \(W^0\)}
\label{subsec: Merging WV and WO}
% We now justify our motivation. \ignacio{I would rather bring the motivation forward (now you have it as a conclusion, and then justify it --ie, reverse the order of the pars. below and start with "Optimized
%   Attention, instead of using two consecutive linear"}
In standard attention, the output \(O\) of the attention layer can be written as
\begin{equation}
  \begin{aligned}
    O & = (H_1 \ \cdots \ H_h) W^O \\
    & = (S_1 V W^V_1 \ \cdots \ S_h V W^V_h)
    \begin{pmatrix}
      W^O_1 \\ \vdots \\ W^O_h
    \end{pmatrix}\\
    & = S_1 V W^V_1 W^O_1 + \cdots + S_h V W^V_h W^O_h,
  \end{aligned}
\end{equation}
where \(W^O_i\) is the matrix %that contains 
with rows
\((i-1) d_v + 1, \dots, i d_v\) of \(W^O\) for \(i = 1, 2, \dots, h\).
By the rank-nullity theorem, for each head, we have that:
\begin{equation*}
  \begin{aligned}
    \dim & (\spn(V W^V_i W^O_i)) \\
    & = \rank(V W^V_i W^O_i) \leq \rank(W^V_iW^O_i), \\
    & \leq \min(\rank(W^V_i), \rank(W^O_i)) \\
    & = \min(d_m, d_v) = d_v.
  \end{aligned}
\end{equation*}
That is, 
\(V W^V_i W^O_i\) has at most \(d_v\) independent
columns, and the linear function \(V \mapsto V W^V_i W^O_i\) maps the
columns of \(V\) into a \(d_v\)-dimensional subspace of \(\RR^{d_m}\).
Thus, standard attention uses two consecutive matrix multiplications to
embed the columns of \(V\) into a \(d_v\)-dimensional subspace of
\(\RR^{d_m}\), which does not align with Principle~\ref{Princ: Linear Combo}.

To address this, in Optimized Attention, we absorb
% We focus on \eqref{eq: Legacy O} and \eqref{eq:
%   Legacy V} in standard attention. We propose absorbing
\(W^V_1, W^V_2, \dots, W^V_h\) into \(W^O\)
 in Eqs.~\eqref{eq: Legacy O} and \eqref{eq:
  Legacy V}, thus reducing the
computational cost of the attention layer by \(h\) matrix
multiplications at a very limited performance cost--which we evaluate
in \cref{sec: Evaluation}.
  
Optimized
Attention uses one slicing and one linear transformation (see \cref{subfig: Optimised Attention} and \cref{def: Merging WV and WO}), instead of the two consecutive linear transformations (one downscaling and one upscaling).
Specifically, instead of multiplying \(V\) from the right by
\(W^V_i\), we slice \(V\) into \(V_1, \dots, V_h\), where
\(V_i\) consists of columns \((i-1) d_v + 1, \dots, i d_v\) of \(V\),
and then, instead of computing \(S_i V W^V_i W^O_i\), we compute
\(S_i V_i W^O_i\), which needs fewer parameters and matrix
multiplications (see \cref{rem: WV and WO} and \cref{subsec: Speed and FLOPS Analysis} for theoretical and empirical evaluations, respectively.)

% \smallskip
%
\begin{definition}[Optimized Attention]
  \label{def: Merging WV and WO}
  Using the notation of \cref{def: Legacy Attention},
  \emph{Optimized Attention} is defined as follows: %with the following equations:
  \vspace{-0.2em}
  \begin{center}
  \begin{tcolorbox}[colback=GreyColour,colframe=black,arc=5pt,boxrule=1pt,width=19em, top=-13pt, left=0pt, bottom=-1pt]
  \begin{align}
    O\  & = (H_1, H_2, \dots, H_h) W^O, \label{eq: Optimised O}\\
    H_i & = S_i V_i, \label{eq: Optimised Y}\\
    S_i & = \softmax(\frac{Q'_i\tr{K'}_i}{\sqrt{d_k}}), \label{eq: Optimised S}\\
    K'_i & = K W^K_i, \label{eq: Optimised K}\\
    Q'_i & = Q W^Q_i. \label{eq: Optimised Q}
  \end{align}
  \end{tcolorbox}
  \end{center}
\end{definition}
% 
% \medskip

%
\begin{remark}
  \label{rem: WV and WO}
  Optimized Attention is more efficient than standard attention, % in that it has 
  having \(h\) matrix multiplication and \(d^2_m\)
  parameters less than standard attention.
\end{remark}
\begin{proof}
  Compared to Optimized Attention, standard attention has extra
  \(W^V_1, W^V_2, \dots, W^V_h\), which are multiplied from the right
  to \(V\). This amounts to a total of \(d_m d_v h = d_m^2\) parameters
  and \(h\) matrix multiplications.
\end{proof}
\subsection{\!\!\!\! Efficient Attention:\! Absorbing \(W^K\)\! into \(W^Q\)\!}
\label{subsec: Merging WQ and WK}
In \cref{subsec: Merging WV and WO}, we discussed our motivation behind dropping
\(W^V\). Here, we do the same for \(W^K\) to further reduce the
computational cost of the attention mechanism. Before this, we note that for the
pre-\(\softmax\) attention scores for each head, we have: % that:
\begin{equation*}
  \begin{aligned}
    & \dim(\spn(\frac{Q W_i^Q \tr{W^K}_i \tr{K}}{d_k}) \\
    & \ = \rank(Q W_i^Q \tr{W^K}_i \tr{K}) \leq \rank(W_i^Q \tr{W^K}_i),\\
    & \ \leq \min(\rank(W_i^Q), \rank(W^K_i)) \\
    & \ = \min(d_m, d_k) = d_k.
  \end{aligned}
\end{equation*}
More precisely, here two linear kernels, \(W_i^Q\) and \(\tr{W^K}_i\), are stacked-- this opposes Principle~\ref{Princ: Linear Combo}. Thus, following the same approach as in Optimized Attention, we merge \(\tr{W^K}_i\) into \(W_i^Q\) and replace the \(W^K_i\) linear transformation by slicing as depicted in \cref{subfig:
  Efficient Attention} and defined in \cref{def: Merging WQ and WK}.

% \smallskip
%
\begin{definition}[Efficient Attention]
  \label{def: Merging WQ and WK}
  Using the same notation as \cref{def: Merging WV and WO},
  we define \emph{Efficient Attention} with the following equations:
  \vspace{-0.2em}
  \begin{center}
  \begin{tcolorbox}[colback=GreyColour,colframe=black,arc=5pt,boxrule=1pt,width=19em, top=-13pt, left=0pt, bottom=-1pt]
  \begin{align}
    O\  & = (H_1, H_2, \dots, H_h) W^O, \label{eq: Efficient O}\\
    H_i & = \ S_i V_i, \label{eq: Efficient H}\\
    S_i & = \ \softmax(\frac{Q'_i \tr{K}_i}{\sqrt{d_k}}), \label{eq: Efficient S}\\
    Q'_i & = \ Q W_i^Q, \label{eq: Efficient Q}
  \end{align}
  \end{tcolorbox}
  \end{center}
  where \(K_i\) denotes the subtensor consisting of
  \((i-1) d_k + 1, \dots, i d_k\) rows from \(K\).
\end{definition}
%
% \medskip

%
\begin{remark}
  \label{rem: WQ and WK}
  Efficient Attention is more efficient than standard and Optimized Attention
  as it has \(h\) matrix
  multiplication and \(d^2_m\) parameters less than Optimized
  Attention and \(2h\) multiplication and \(2d_m^2\) parameters fewer
  than standard attention.
\end{remark}
\begin{proof}
  In Efficient Attention, we do not have
  \(W^K_1, W^K_2, \dots, W^K_h\), which are applied to \(K\) from
  left. Hence, we reduce the
  number of matrix multiplications by \(h\) and parameters by
  \(d_m^2\), compared to Optimized Attention. From this and \cref{rem: WV and WO}, it follows that
  Efficient Attention has \(h + h = 2h\) matrix multiplication and
  \(d_m^2 + d_m^2 = 2 d_m^2\) parameters fewer than standard attention.
\end{proof}
\subsection{Super Attention: Introducing \(W^A\)}
\label{subsec: Introducing WA}
Looking at the Eqs.~(\ref{eq: Legacy O}-\ref{eq: Legacy Q}),
we observe that in SDPA, there are learnable parameters between \(Q\) and \(K\);
however, there is no such parameter between \(K\) and \(V\) (even though a \(\softmax\) is applied to the term containing \(K\)). Following Principle 2, we introduce a new learnable parameter \(W^A\) that linearly transforms the values from the left.
To better observe this, let us write the equation for one
head in one of the attention variants, e.g.,
Efficient Attention by combining Eqs.~(\ref{eq: Efficient
  H}--\ref{eq: Efficient Q}):
\begin{equation}
  \label{eq: Efficient Attention}
  H_i = \softmax(\frac{Q W_i^Q \tr{K}_i}{d_m}) V_i W^O.
\end{equation}
As we see in \cref{eq: Efficient Attention}, there are no learnable
parameters between \(\tr{K}\) and \(V\), and the attention scores
\(S_i\) are directly applied to the values \(V_i\). The intuition
behind directly applying \(S_i\) to \(V_i\) is that the attention
scores in \(S_i\) determine ``how much attention is paid'' to each of
the features of each token in \(V_i\). Despite this intuition, we
found that in practice the model can benefit from an additional kernel which comes in between the scores \(S_i\) and values \(V_i\). Specifically, with the introduction of
\(W^A\), \cref{eq: Efficient Attention} changes to
\begin{equation}
  \label{eq: Super Attention}
  H_i = \softmax(\frac{Q W_i^Q \tr{K}_i}{d_m}) W^A V_i W^O.
\end{equation}

The role of \(W^A\) is to mix and align the values vertically
(token-wise). Thus, to prevent ``look ahead'' in the attention
mechanism for use in causal language modelling, we can constrain
\(W^A\) to be lower triangular, so that future tokens do not influence
the current one in \(W^A\). Note that we use the same \(W^A\) for all
heads. The reason here is that we want to improve the model
performance while keeping the model size as small as possible. Thus,
in a more general formulation, one can use different \(W^A\) for each
head to perhaps gain even better performance, but at the cost of increasing
the number of parameters, and thereby the model size.

% \smallskip
%
\begin{definition}[Super Attention]
  \label{def: Introducing WA}
  Using the notation of \cref{def: Merging WQ and WK},
  we define \emph{Super Attention} with the 
  following equations:
  \vspace{-0.2em}
  \begin{center}
  \begin{tcolorbox}[colback=GreyColour,colframe=black,arc=5pt,boxrule=1pt,width=19em, top=-13pt, left=0pt, bottom=-1pt]

  \begin{align}
    O\  & = (H_1, H_2, \dots, H_h) W^O, \label{eq: Super O}\\
    H_i & = \ S_i V'_i, \label{eq: Super H}\\
    S_i & = \ \softmax(\frac{Q'_i \tr{K}_i}{\sqrt{d_k}}), \label{eq: Super S}\\
    V_i' & = \ W^A V_i,\label{eq: Super V}\\
    Q_i' & = \ Q W_i^Q, \label{eq: Super Q}
  \end{align}
  \end{tcolorbox}
  \end{center}
  where \(W^A \in \RR^{\ell \times \ell}\) is the \emph{alignment
    kernel}, which vertically (i.e., for values corresponding to
  different tokens) aligns and mixes the values before the attention
  scores are applied to them.
\end{definition}
%
% \medskip

%
\begin{remark}
  \label{rem: WA}
  Super Attention is more efficient than standard attention whenever
  the model dimension \(d_m\) is greater than or equal to the context
  length \(\ell\). This means that Super Attention has at least \(h\)
  matrix multiplication and \(d_m^2\) parameters fewer than standard
  attention.
\end{remark}
\begin{proof}
  Looking at Eqs. (\ref{eq: Efficient O}--\ref{eq: Efficient
    Q}) and (\ref{eq: Super O}--\ref{eq: Super Q}),
  Super and Efficient Attention have the same
  equations, except that Super Attention has an additional linear
  transformation in \cref{eq: Super V}, where \(V_i\)'s are multiplied
  by \(W^A\) from the left. This amounts to \(\ell^2\) parameters and
  \(h\) matrix multiplication more than Efficient Attention. From
  \cref{rem: WQ and WK}, it follows that Super Attention has at least
  \(2 d_m^2 - \ell^2 \geq d_m^2\) parameters and \(2h -h = h \) matrix
  multiplications fewer than standard attention.
\end{proof}
%

%%% Local Variables:
%%% mode: latex
%%% TeX-master: "../main"
%%% End:

%
%%
\section{Evaluation}
\label{sec: Evaluation}
We evaluate the proposed mechanism on a range of NLP tasks (\cref{subsubsec: NLP,subsec: LLM}); we then show that the approach generalises to other modalities by evaluating them on a number of vision benchmarks (\cref{subsec:
  Vision}). We also provided a detailed
comparison of the computational costs and edge device performance in
\S~\ref{subsec: Speed and FLOPS Analysis}, \ref{subsubsec: Edge Device}, and \ref{app: FLOPs}.

\vspace{-0.2em}
\paragraph{Evaluation Methodology.} We evaluate on a range of benchmarks. In each benchmark, we follow the common practices for evaluating the performances. For all
benchmarks, (1) we use the same model architecture and iterate
between standard, Optimized, Efficient, and Super Attention; (2) we
continue training until validation loss flattens or a given computational budget is reached; and (3) for benchmarks on smaller datasets, we report the results by averaging over five runs to ensure fairness.

\vspace{-0.2em}
\paragraph{Experimental Setup.} All experiments in \cref{subsec: Vision,subsubsec: NLP} are implemented in
Keras with JAX backend using 
\href{https://keras.io/examples}{\texttt{keras.io/examples}} with
minor dataset-specific adjustments, e.g., modifying the number of
classes, layers, etc. The generative language modelling experiment in \cref{subsubsec: NLP} is an adaptation of Andrej Karpathy's NanoGPT \citep{Karpathy22}.
All the reported results are obtained by
training on an Nvidia RTX 4090 GPU (24GB VRAM) or an Nvidia A100 GPU
(80GB VRAM); however, we have chosen model and batch sizes to ensure
that they run on 24GB VRAM. In each table, we report the train and
test loss and  accuracy (where relevant), the number of parameters in one attention
layer (in the ``\# Param.'' column), the average training time (in
seconds) of models for one epoch on an RTX 4090 GPU (in the ``Epoch
Time'' column), as well as other related task-specific metrics.

% Precisely,
% Super Attention 
% \(\nicefrac{(60 - 51.18)}{51.18} =17.23\%\) on ImageNet) despite
% having fewer parameters and being faster to train (up to
% \(\nicefrac{(8.31 - 7.58)}{7.58} =5.86\%\) on MNIST). Efficient and
% Optimized Attention perform comparably to standard attention in terms
% of accuracy and loss but with \(\nicefrac{1}{2}\) and
% \(\nicefrac{3}{4}\) as many attention parameters, respectively. They
% are also significantly faster to train (up to
% \(\nicefrac{(8.31 - 7.05)}{8.31} = 15.16\%\) when comparing Efficient
% Attention to standard attention on MNIST).

%
%%
\vspace{-0.2em}
\subsection{NLP Benchmarks}
\vspace{-0.2em}
\label{subsubsec: NLP}
In this section, we evaluate the attention variants 
in Transformer models of different scales for three NLP tasks:
sentiment classification, Machine Translation (MT) and
generative language modelling (LM) and NLI tasks. 

\vspace{-0.2em}
\paragraph{Sentiment Classification.} For sentiment
classification (\cref{tbl: NLP}), we use two widely-used benchmarks, IMDB Movie Reviews 
\citep{Maas+11} and Amazon Reviews \citep{NiM19} datasets. The dataset sizes for these two experiments in this part are 50k and 3.65M, and the model sizes are ~650K and ~26M parameters, respectively.

\vspace{-0.2em}
\paragraph{Machine Translation (MT)} For MT (\cref{tbl: Translation}), we
use the combined Europarl \citep{Koehn05} and Anki \citep{anki}
dataset for English-to-Spanish translation. The dataset includes ~2 million pairs and the model sizes range from ~93-104 million parameters for different architectures.

\vspace{-0.2em}
\paragraph{Generative LM and NLI.} For generative language modelling (\cref{tbl: Language Modelling}),
we use the OpenWebText dataset \citep{GokaslanC19} for training and the HellaSwag dataset \citep{Zellers+19} for comparing the common-sense reasoning performance of the trained models. This dataset includes more than 9 Billion tokens and the model sizes range between 110-124 million parameters for different architectures. The context window of the language models is set to 1024 tokens.

\begin{table*}[th]
  \centering
  \caption{Sentiment classification results, averaging 
  over five runs 
    on IMDB and Amazon Reviews
    datasets. Numbers in parentheses indicate the ranking of each
    attention variant for a given metric and dataset. Ablation studies on
    the number of heads for all experiments is available in \cref{app:
      NLP}.  Efficient Attention models
    have the smallest attention layer size and the Super Attention
    models perform the best in terms of accuracy and loss.}
  \label{tbl: NLP}
  \vspace{-0.4em}
  \resizebox{0.9\textwidth}{!}{%
  \begin{tabular}{clrrrccccc}
    \toprule
    Dataset & Att. & $h$ & $d_m$ & \# Param. & Epoch Time & Acc. (\%) & Loss  & Val Acc. (\%) & Val Loss\\
    \midrule

    % \rowcolor{StnColour}
     & Stn. & 4 & 32 & 4,224 (4) & 0.315 (4) & 95.70 (4) & 0.086 (3) & 77.62 (4) & 0.474 (4) \\

     % \rowcolor{OptColour}
     & Opt. & 4 & 32 & 3,168 (2) & 0.305 (3) & 96.31 (3) & 0.095 (4) & 77.85 (2) & 0.472 (2) \\

    % \rowcolor{EffColour}
    & Eff. & 4 & 32 & \textbf{2,112 (1)} & \textbf{0.280 (1)} & 96.41 (2) & \textbf{0.064 (1)} & 77.77 (3) & \textbf{0.468 (1)} \\
    
    % \rowcolor{SupColour}
    \multirow{-4}{*}{\rotatebox[origin=c]{90}{IMDB}} & Sup. & 4 & 32 & 3,168 (2) & 0.299 (2) & \textbf{97.45 (1)} & 0.070 (2) & \textbf{78.34 (1)} & 0.472 (2)\\
       
    \midrule

    % \rowcolor{StnColour}
     & Stn. & 4 & 128 & 66,048 (4) & 66.97 (4) & 88.49 (3) & 0.25 (3) & 65.55 (4) & 0.77 (3)\\

    % \rowcolor{OptColour}
     & Opt. & 4 & 128 & 49,536 (3) & 61.75 (3) & \textbf{89.56 (1)} & \textbf{0.23 (1)} & 65.67 (2) & 0.75 (2) \\

    % \rowcolor{EffColour}
     & Eff. & 4 & 128 & \textbf{33,024 (1)} & \textbf{56.44 (1)} & 86.63 (4) &  0.29 (4) & 65.58 (3) & 0.77 (3) \\

     % \rowcolor{SupColour}
    \multirow{-4}{*}{\rotatebox[origin=c]{90}{Amazon}} & Sup. & 4 & 128 & 42,336 (2) & 59.86 (2) & 88.56 (2) & 0.24 (2)  & \textbf{68.10 (1)} & \textbf{0.71 (1)} \\
    
    \bottomrule
  \end{tabular}%
  }
\end{table*}
\begin{table*}[th]
  \centering
  \caption{Machine translation results, averaging
  over five runs for
    English-to-Spanish MT on combined Europarl and Anki translation
    datasets. Numbers in parentheses indicate the ranking of each
    attention variant for that metric. Ablation 
    on the number of heads is available in \cref{app: NLP}. Optimized and Efficient Attentions perform similarly to
    standard attention on most metrics with \nicefrac{1}{2} and
    \nicefrac{3}{4} as many attention parameters, respectively. As the Super Attention layer has a fixed context length and the decoder requires a varying context length, using Super Attention would require using a sliding window, which would not be comparable to the full attention used for the other variants.}
  \label{tbl: Translation}
  \vspace{-0.4em}
  \resizebox{\textwidth}{!}{%
  \begin{tabular}{lrrrrccccccc}
    \toprule
    Att. & $h$ & $d_m$ & $d_k$ & \# Param. & Epoch Time & BLEU & Acc. & Loss & Val BLEU & Val Acc. & Val Loss\\
    \midrule
    % \rowcolor{StnColour}
        Stn. & 4 & 1024 & 256 & 4.2M (3) & 600.0 (3) & 23.1 (2) & 81.11 (3) & 0.83 (3) & \textbf{22.8 (1)} & 81.41 (3) & 0.84 (3) \\
    % \midrule
    % \rowcolor{OptColour}
        Opt. & 4 & 1024 & 256 & 3.1M (2) & 586.8 (2) & \textbf{24.5 (1)} & \textbf{82.06 (1)} & \textbf{0.78 (1)} & 22.6 (3) & \textbf{81.98 (1)} & \textbf{0.80 (1)} \\
    % \midrule
    % \rowcolor{EffColour}
    Eff. & 4 & 1024 & 256 & \textbf{2.1M (1)} & \textbf{523.0 (1)}  & 22.6 (3) & 81.15 (2) & 0.82 (2) & 22.3 (3) & 81.44 (2) & 0.83 (2) \\
    \bottomrule
  \end{tabular}%
  }
  \vspace{-0.8em}
\end{table*}

\vspace{-0.2em}
\paragraph{NLP Results Analysis.} Super
Attention outperforms attention variants in terms of
validation accuracy (up to \(\nicefrac{(68.10 - 65.55)}{65.55} = 3.89\%\)
compared to standard attention on Amazon Reviews) in the sentiment classification task. Similarly, we see for MT as well as generative LM and NLI tasks that the Optimized and Efficient architectures perform closely or on par with the Standard mechanisms. We also observe
that standard attention is slower than all other variants
(up to \(\nicefrac{(600 - 523)}{523} = 14.72\%\) slower than Efficient
Attention in MT) with the highest number of parameters (twice as many
parameters per layer compared to Efficient Attention). The generative LM experiment reveals subtle differences in performance among the models in training performance; However, our NLI experiment shows that when evaluated on the HellaSwag benchmark, all three models exhibit comparable performance, achieving accuracy rates between 30\% and 31\%.

\begin{table*}[th]
  \vspace{-0.2em}
  \centering
  \caption{Averages of different metrics in generative LM using NanoGPT, a widely-referenced re-implementation of GPT-2 124M by Andrej Karpathy, based on different attention variants. The models are trained on the OpenWebText dataset ($\sim$9B training tokens) for one epoch with a batch size of 500 and a micro-batch size of 5 using a single A100 80GB node. The context window is 1024. In addition to the loss and perplexity, we provide the size of each model and the result of NLI on the HellaSwag benchmark. Similarly to the MT task, a fair comparison of Super Att. against other variants is not feasible as NanoGPT uses full attention but Super Att. requires using a sliding window.}
  \label{tbl: Language Modelling}
  \vspace{-0.3em}
  \resizebox{0.95\textwidth}{!}{%
  \begin{tabular}{lrrrccccccc}
    \toprule
    Att. & $h$ & $d_m$ & $d_k$ & Layer Size & Model Size &Train Loss & Train PPL & Val Loss & Val PPL & HellaSwag \\
    \midrule
    % \rowcolor{StnColour}
        Stn. & 12 & 768 & 64 & 2.36M & 124M & 2.92 & 18.5 & 3.13 & 22.9 & 0.31 \\ 
    % \rowcolor{OptColour}
        Opt. & 12 & 768 & 64 & 1.77M & 117M & 2.96 & 19.3 & 3.14 & 23.1 & 0.31 \\
    % \rowcolor{EffColour}
    Eff. & 12 & 768 & 64 & 1.18M & 110M & 3.02 & 20.5 & 3.18 & 24.0 & 0.30 \\
    \bottomrule
  \end{tabular}%
  }
\end{table*}
\subsection{Vision Transformers}
\label{subsec: Vision}
\vspace{-0.3em}
 We experiment with three widely adopted vision datasets of varying size and complexity: MNIST \citep{Lecun+MNIST},
CIFAR100 \citep{Krizhevsky09}, and ImageNet1K
\cite{Russakovsky+15}. For Brevity, we refer to the ImageNet1K dataset
throughout the paper as ImageNet. Note that for the reported ImageNet results in \cref{tbl: Vision}, we first pre-trained the model on the ImageNet21K dataset. We report the training details in \cref{app: Vision}.

\begin{table*}[th]
  \centering
  \caption{Vision results, averaging over five runs on
    MNIST and CIFAR100, and one run on ImageNet. Numbers in parentheses
    indicate the ranking of each mechanism for a given metric and dataset. An ablation study on the number of heads is available in \cref{app: Vision}. An additional
    ablation study for models of the same size on ImageNet but with different
    attention mechanisms is provided in \cref{app: Vision}. As
    expected, Efficient Attention models have the
    smallest attention layer size, and the Super Attention models
    achieve the highest accuracy and lowest loss.}
  \label{tbl: Vision}
  \vspace{-0.3em}
  \resizebox{\textwidth}{!}{%
  \begin{tabular}{clrrrccccccc}
    \toprule
    Dataset & Att. & $h$ & $d_m$ & \# Param. & Epoch Time & Acc. (\%) & Loss & Top 5 & Val Acc. (\%) & Val Loss & Val Top 5\\
    \midrule

    % \rowcolor{StnColour}
      & Stn. & 4 & 128  & 66K (4) & 8.31 (4) & 93.73 (4) & 0.209 (4) & N/A & 98.12 (4) & 0.062 (4) & N/A \\

    % \rowcolor{OptColour}
      & Opt. & 4 & 128  & 49K (3) & 7.68 (3) & 95.36 (2) & 0.161 (2) & N/A & 98.43 (2) & 0.046 (2) & N/A \\

    % \rowcolor{EffColour}
       & Eff. & 4 & 128  & \textbf{33K (1)}  & \textbf{7.05 (1)} & 94.28 (3) & 0.197 (3) & N/A & 98.27 (3) & 0.058 (3) & N/A\\
    % \rowcolor{SupColour}
       \multirow{-4}{*}{\rotatebox[origin=c]{90}{MNIST}} & Sup. & 4 & 128 & 37K (2) & 7.58 (2) & \textbf{96.96 (1)} & \textbf{0.112 (1)}  & N/A & \textbf{98.62 (1)} & \textbf{0.051 (1)} & N/A \\
       
    \midrule

    % \rowcolor{StnColour}
     & Stn. & 8 & 256  & 263K (4) & 21.19 (4) & 72.28 (2) & 1.41 (2) & 91.02 (2) & 48.14 (3) & 1.82 (3) & 90.22 (4) \\

     % \rowcolor{OptColour}
     & Opt. & 8 & 256 & 197K (2) & 20.39 (3) & 72.26 (3) & 1.47 (3)  & 93.01 (3) & 48.63 (2)  & 1.71 (2)  & 90.99 (2)  \\

     % \rowcolor{EffColour}
     & Eff. & 8 & 256 & \textbf{131K (1)} & \textbf{19.22 (1)} & 71.96 (4) & 1.49 (4) & 92.23 (4) & 47.95 (4) & 1.83 (4) & 90.48 (3) \\

     % \rowcolor{SupColour}
    \multirow{-4}{*}{\rotatebox[origin=c]{90}{CIFAR100}} & Sup. & 8 & 256  & 197K (3) & 20.28 (2) & \textbf{79.62(1)} & \textbf{1.28 (1)} & \textbf{94.34 (1)} & \textbf{49.28 (1)} & \textbf{1.55 (1)} & \textbf{91.69 (1)} \\

    \midrule

    % \rowcolor{StnColour}
     & Stn. & 12 & 768  & 2.36M (4) & 2572 (4) & 92.07 (2) & 1.02 (2) & 98.41 (2) & 74.35 (3) & 1.47 (3) & 94.10 (4) \\
     
    % \rowcolor{OptColour}
    & Opt.  & 12 & 768 & 1.77M (3) & 2426 (2) & 91.78 (3) & 1.03 (3) & 98.36 (3) & 77.12 (2) & 1.47 (3) & 94.21 (3) \\
        
    % \rowcolor{EffColour}
    & Eff. & 12 & 768 &\textbf{ 1.18M (1)}  & \textbf{2374 (1)} & 90.36 (4) & 1.05 (4) & 98.37 (4) & 75.67 (4) & 1.44 (2) & 95.46 (2) \\
    
    % \rowcolor{SupColour}
    \multirow{-4}{*}{\rotatebox[origin=c]{90}{ImageNet}} & Sup. & 12 & 768 & 1.22M (2) & 2483 (3) & \textbf{94.09 (1)} & \textbf{0.94 (1)} & \textbf{99.32 (1)} & \textbf{79.29 (1)} & \textbf{1.39 (1)} & \textbf{96.37 (1)} \\
    
    \bottomrule
  \end{tabular}%
  }
  \vspace{-0.9em}
\end{table*}

\paragraph{ViT Results Analysis.} The number of parameters in the models
considered for the vision tasks range from 300K (MNIST) to 60M
(ImageNet), their context length ranges from 64 (MNIST) to 256
(CIFAR100 and ImageNet), the dataset sizes range from 60K (MNIST) to
1.28M (ImageNet), and the number of classes ranges from 10 (MNIST) to
1K (ImageNet). Similar to text Transformers, ViTs using Super Attention architecture perform better than all other variants despite having fewer
parameters than standard attention. Also, Optimized and Efficient Attentions perform
 comparably to standard attention with fewer parameters.

\subsection{Speed and FLOPs Analysis}
\label{subsec: Speed and FLOPS Analysis}
\cref{subsubsec: Edge Device,app: FLOPs} are dedicated to studying the computational complexity and inference speed of the considered attention variants. \cref{eq: FLOPs Formula} formulates the computational complexity for each algorithm.

\begin{figure}[th]
  \centering
  \begin{subfigure}[b]{0.49\columnwidth}
    \centering
    \includegraphics[width=\columnwidth]{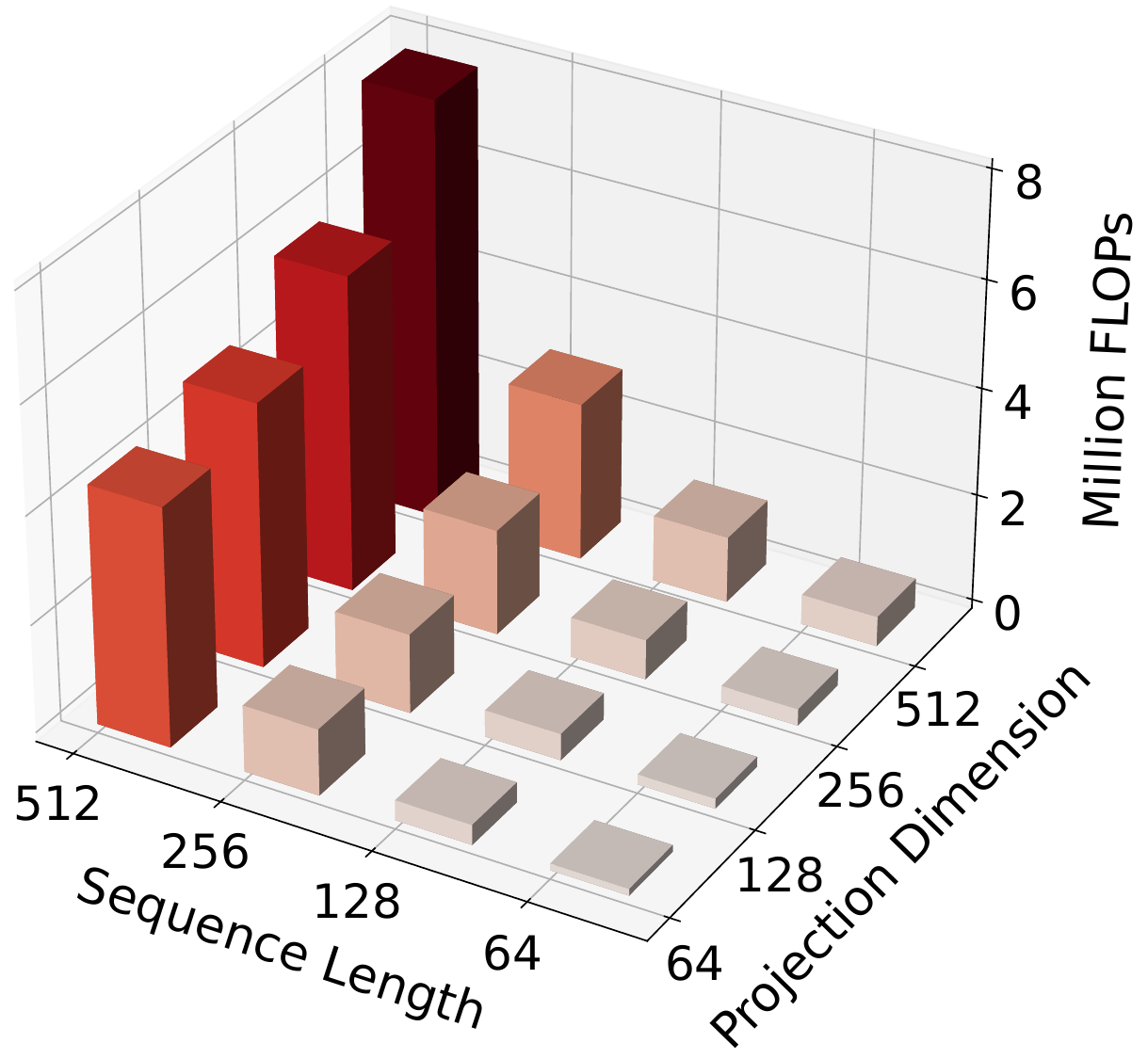}
    \subcaption{Standard Att.}
  \end{subfigure}
  %
  % \hfill
  % %
  % \begin{subfigure}[b]{0.24\textwidth}
  %   \centering
  %   \includegraphics[width=\textwidth]{flops_single_head/OptAttn_single_head}
  %   \subcaption{Optimized Att.}
  % \end{subfigure}
  % %
  \hfill
  \begin{subfigure}[b]{0.49\columnwidth}
    \centering
    \includegraphics[width=\columnwidth]{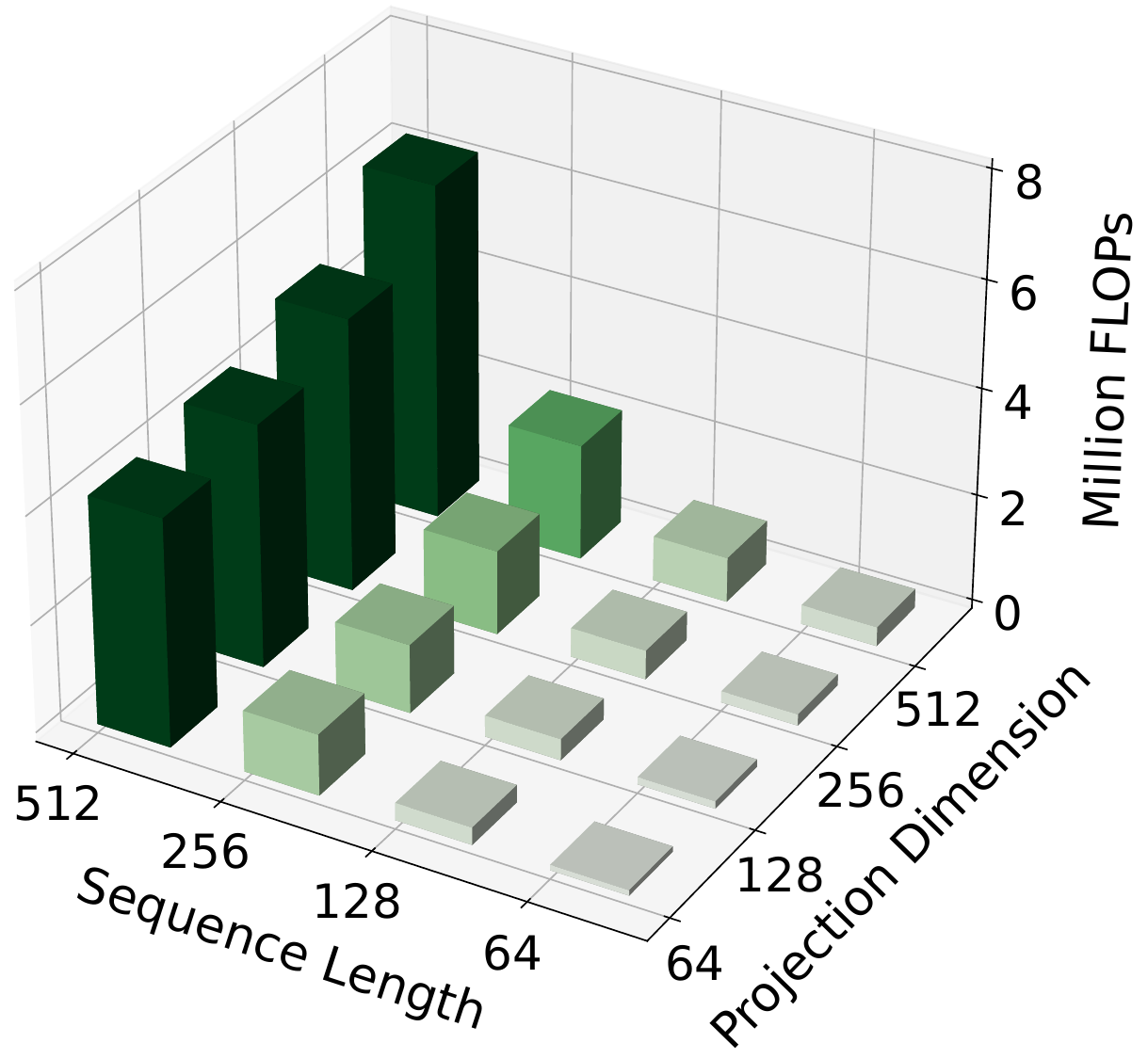}
    \subcaption{Efficient Att.}
  \end{subfigure}
  %
  % \hfill
  % %
  % \begin{subfigure}[b]{0.24\textwidth}
  %   \centering
  %   \includegraphics[width=\textwidth]{flops_single_head/SupAttn_single_head}
  %   \subcaption{Super Att.}
  % \end{subfigure}
  % %
  \vspace{-1.2em}
  \caption{3D plots visualizing the number of FLOPs for a forward + backward pass given different sequence lengths
    and projection dimensions in single-head setting for Efficient and Standard attention. Efficient Att. needs substantially fewer FLOPs for completing
    a forward + backward pass. \cref{fig: Flops All Archs Full} compares all architectures.}
  \label{fig: Flops All Archs}
  \vspace{-1.3em}
\end{figure}

\begin{figure}[t!]
  \centering
  \includegraphics[width=\columnwidth]{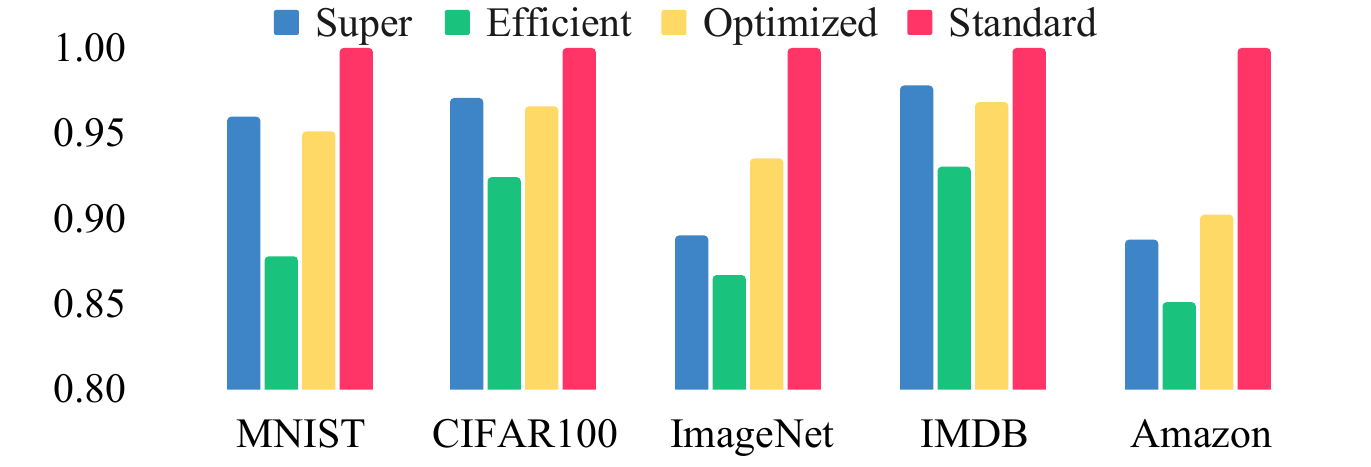}
  \vspace{-1.5em}
  \caption{Summary of relative inference latency of models using different attention variants relative to standard attention on different datasets on Edge Device (Apple Laptop M2). Efficient Att. is the fastest (Optimized and Super Att. are also faster than standard attention). More details and numerical results are available in \cref{tbl: inference time}.}
  \label{fig: FLOPs EdgeDevice}
   \vspace{-1.3em}
\end{figure}

\cref{fig: Flops All Archs,fig: Flops All Archs Full}  visualize a comparison between the required FLOPs for each algorithm based on ``sequence length'' and ``projection dimension''. It indicates Efficient Attention requires the least number of FLOPs under all scenarios. From an empirical perspective, \cref{tbl: inference time} and \cref{fig: FLOPs EdgeDevice} exhibit the faster inference speed (lower latency) of Efficient Attention compared to other variants in all datasets, followed by Optimized and Super variants.

\begin{figure*}[h!]
  \centering
  \begin{subfigure}[b]{\textwidth}
    \centering
    \includegraphics[width=0.5\textwidth]{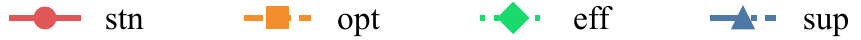}
  \end{subfigure}
  \hfill
  \begin{subfigure}[b]{\textwidth}
    \centering
    \includegraphics[width=\textwidth]{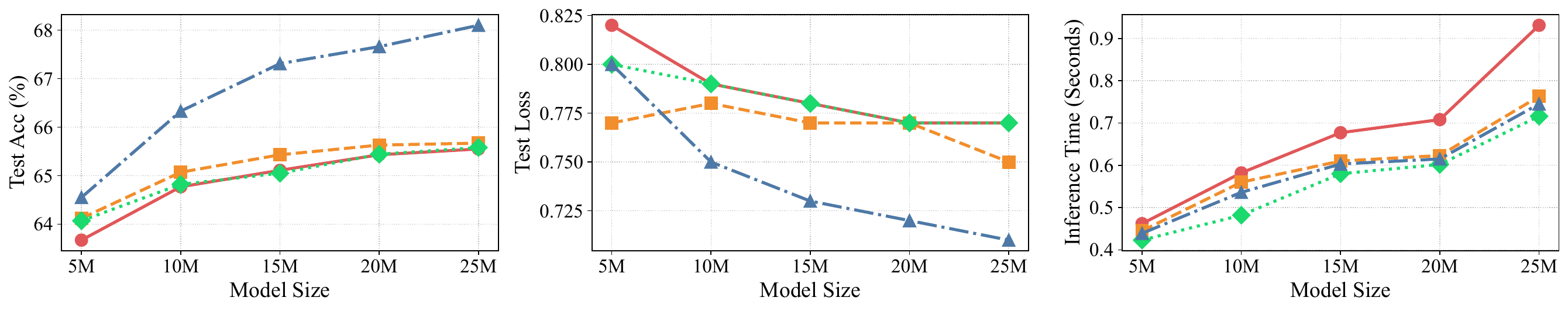}
  \end{subfigure}
  \vspace{-1.7em}
  \caption{Performance of different architectures on the Amazon Reviews as the size of models grows from 5 Million parameters to 25 Million parameters. In terms of test accuracy and loss, Super Attention shows increasingly better performance compared to all other architectures which are performing on par with each other. In terms of inference speed, all variants (especially Efficient) perform better than the Standard attention.}
  \label{fig: Scaling Test}
  \vspace{-0.9em}
\end{figure*}
\subsection{Scaling Analysis}
\label{subsec: Scaling Analysis}
\vspace{-0.3em}
We analyzed scaling behaviour across three dimensions: attention heads, dataset size, and model size. Head scaling experiments across tasks (\cref{tbl: inference time,tbl: MNIST,tbl: IMDB,tbl: Amazon,tbl: Translation Complete}) showed consistent performance improvements with increased heads for all architectures. Dataset scaling ranged from IMDB (50K examples) to OpenWebText (9B tokens) for language tasks, and MNIST (60K examples) to ImageNet (1.28M examples) for vision tasks, with our variants maintaining their relative performance advantages across scales. Model scaling experiments on Amazon Reviews (\cref{fig: Scaling Test,fig: Scaling All}) demonstrate that as models grow from 5M to 25M parameters, Super Attention consistently outperforms standard attention, while Optimized and Efficient variants match standard's performance with significantly fewer parameters. Notably, standard attention's computational inefficiency becomes more pronounced at larger scales in both training and inference.

%%% Local Variables:
%%% mode: latex
%%% TeX-master: "../main"
%%% End:

%
%%
\vspace{-0.2em}
\section{Related Work}
\label{sec: Related Work}
\vspace{-0.4em}
Since their adoption, many research directions have emerged to address various shortcomings of attention mechanisms and Transformer models. Sparse attention, such as Longformer
\citep{BeltagyPC20, Zhang+21longformer}, reduces the computational
complexity by focusing on key input parts \citep{ChildGRS19}. Despite handling long sequences efficiently, sparse
mechanisms struggle with tasks requiring a comprehensive sequence
analysis.

%\textcolor{red}{
Another line of research focuses on approximating the attention matrix to attain linear complexity. Performer \citep{Choromanski+21} uses random feature maps and FAVOR+ mechanism; Linformer \citep{Wang+20} projects keys and values to lower dimensions by exploiting low-rank properties. While these approaches achieve efficiency through approximation, they often compromise model quality. In contrast, our proposed variants achieve efficiency through structural modifications while maintaining or improving model quality.
%}

Recent work has explored architectures that combine transformers' parallel training capabilities with RNNs' inference efficiency, including RWKV \citep{Peng+23} with linear recurrence and State-Space models like Mamba \citep{GuD24} and S4 \citep{GuGKR21}. While these approaches show promise, they require fundamental architectural changes. Our work instead focuses on optimizing the attention mechanism itself, preserving the proven benefits and versatility of transformer architectures while reducing computational costs.

Several approaches focus on reducing model redundancy. \citet{Voita+19} demonstrate that multi-head SDPA is over-parameterized, leading to collaborative frameworks that reduce projection sizes \citep{CordonnierLJ20}. Similarly, sparsification techniques reduce non-zero elements in weights, with recent work achieving 1-10\% compression with minimal performance impact \citep{Ashkboos+24}, though potentially affecting robustness \citep{Timpl+22}. While these approaches focus on post-hoc optimization or pruning, our work fundamentally reimagines the attention mechanism's structure to achieve efficiency by design. We discuss further related attempts (including LoRA, Quantization and Flash Attention) for facilitating the deployability of transformers in \cref{app: Related Work 2}.

%%% Local Variables:
%%% mode: latex
%%% TeX-master: "../main"
%%% End:

%
%%
\vspace{-0.2em}
\section{Discussion and Conclusions}
\vspace{-0.2em}
We proposed and evaluated three variants of SDPA that alter the standard arrangement of linear transformations to achieve better performance per computation cost and number of parameters (see \cref{fig: Attention Flowcharts} for visualizations). Optimized and Efficient Attention replace one (values) and two (values and keys) linear transformations with slicing, resulting in 25\% and 50\% size reductions and fewer matrix multiplications, respectively. The third variant, Super Attention, introduces a new linear transformation operating on the values from the left. While Super Attention can be applied to standard, Optimized, or Efficient Attention, we combined it with Efficient Attention, resulting in approximately 25\% fewer parameters compared to standard attention.

Our evaluation spanned a wide range of tasks, including sentiment classification on IMDB and Amazon Reviews, Machine Translation on combined Europarl and Anki datasets, generative LM on OpenWebText dataset and NLI on HellaSwag. We used benchmarks varying in size from 50,000 examples to 9 billion tokens. To verify if these architectural benefits generalize across modalities, we also evaluated all variants for image classification on MNIST, CIFAR100, and ImageNet1K.

The experimental results demonstrate that Optimized and Efficient Attention performed comparably to standard attention across different benchmarks, despite having 25-50\% fewer parameters and being faster. Super Attention consistently outperformed standard variant in all applicable benchmarks, achieving improvements of up to 10\% on CIFAR100 and 4\% on Amazon, while maintaining fewer parameters and faster training and inference.

Our generative LM experiment using a 1.1B Llama-based model in \cref{subsec: LLM} provides insight into these variants' performance at larger scales. Yet realizing their true potential requires evaluation at even larger scales, which are beyond our computational resources. The promising results suggest these attention variants could open new pathways for training and deploying capable models on devices with limited computational resources, like smartphones and small personal devices.

\section*{Limitations}
There are two limitations in this paper. First, Super Attention
supports fixed context length due to the fixed size of
\(W^A\) (see \cref{eq: Super V} and \cref{subfig: Super
  Attention}). Nonetheless, these do not affect the advantages of
Super Attention in many SotA applications such as in ViT. Moreover, this
can be addressed using a sliding window, which is a future work currently
in progress. Second, because of limited computational resources, we
could only validate our hypotheses on models with up to 124 million (1.1 billion considering the language model trained in \cref{subsec: LLM})
parameters trained on datasets with up to 9 billion (30 billion considering \cref{subsec: LLM}) tokens. Further scaling the experiments beyond our computational
resources and training large multi-modal and language models using the
proposed mechanisms could facilitate a better understanding of their performance on industrial scales.

%%% Local Variables:
%%% mode: latex
%%% TeX-master: "../main"
%%% End:

\bibliography{bibliography}

%%%%%%%%%%%%%%%%%%%%%%%%%%%%%%%%%%%%%%%%%%%%%%%%%%%%%%%%%%%%%%%%%%%%%%%%%%%%%%%
%%%%%%%%%%%%%%%%%%%%%%%%%%%%%%%%%%%%%%%%%%%%%%%%%%%%%%%%%%%%%%%%%%%%%%%%%%%%%%%
% APPENDIX
%%%%%%%%%%%%%%%%%%%%%%%%%%%%%%%%%%%%%%%%%%%%%%%%%%%%%%%%%%%%%%%%%%%%%%%%%%%%%%%
%%%%%%%%%%%%%%%%%%%%%%%%%%%%%%%%%%%%%%%%%%%%%%%%%%%%%%%%%%%%%%%%%%%%%%%%%%%%%%%
\newpage
\appendix
% \onecolumn
% \section{Societal Impact}
% This paper has the potential to positively impact society by enabling
% the training of smaller and more efficient models. This can help make progress
% in the democratization of AI by making it more widely accessible to a
% broader part of society. By reducing the computational resources and
% energy consumption required to train and deploy state-of-the-art
% language models and foundational models, our work can help lower the
% barriers to entry and promote widespread access to more powerful
% models.

% Furthermore, the increased efficiency of our proposed methods can lead
% to a reduction in the carbon footprint associated with AI development
% and deployment, aligning with the growing need for more sustainable
% and environmentally friendly technologies. As a result, our research
% has the potential to not only advance the field of AI but also to
% contribute to a more inclusive and sustainable future for AI-driven
% technologies.

%
%%
\section{Reproducibility Statement}
The code for all experiments is provided in the supplementary materials. Publicly available datasets are used, with automatic downloads included in the code, except for the Amazon dataset (link in README). The NanoGPT repository (linked in Experimental Setup) details the generative language modelling experiment. Further implementation details are in Section \cref{sec: Evaluation} and \cref{app: Vision,app: NLP}.

\section{Additional Experiments}

\subsection{Edge Device Performance}
\label{subsubsec: Edge Device}
Our main motivation for introducing Optimized, Efficient, and Super
Attention is to allow running more capable models on edge devices. We
calculated the inference times of the Transformer models, we trained
before, on a MacBook Pro with an M2 Chip for each task/attention
mechanism in \cref{tbl: inference time}. As expected, Efficient models
are the fastest. Also, Super Attention and Optimized Attention models
are faster than their standard counterparts with the same number of
heads while performing equally well as we discussed before.
%
% \begin{wraptable}{r}{22em}
\begin{table*}[h]
  \centering
  \caption{Total inference times (in seconds) for each attention
    mechanism/dataset pair on an Apple M2 chip over 5,000 samples.}
  \label{tbl: inference time}
  \medskip
  \resizebox{0.72\textwidth}{!}{%
  \begin{tabular}{lcccccc}
  \toprule
    Name     & \(h\) & MNIST & CIFAR100 & ImageNet & IMDB & Amazon\\
    \midrule
    % \rowcolor{StnColour}
     & 1 & 4.43 \lflush{} & 34.84 \lflush{} & 299.26 \lflush{} & 0.114 \lflush{} & 0.53 \lflush{} \\
    % \rowcolor{StnColour}
     Standard & 4 & 5.27 \lflush{} & 46.06 \lflush{} & 323.84 \lflush{} & 0.183 \lflush{} & 0.87 \lflush{} \\
    % \rowcolor{StnColour}
    & 8 & 6.89 (4) & 62.08 (4) & 341.69 (4) & 0.266 (4) & 1.34 (4) \\
    \midrule
    % \rowcolor{OptColour}
     & 1 & 4.19 \lflush{} & 33.36 \lflush{} & 281.14 \lflush{} & 0.109 \lflush{} & 0.47\lflush{} \\
    % \rowcolor{OptColour}
     Optimized & 4 & 5.22 \lflush{} & 44.17 \lflush{} & 301.30 \lflush{} & 0.176 \lflush{} & 0.76 \lflush{} \\
    % \rowcolor{OptColour}
     & 8 & 6.37 (2) & 60.63 (2) & 320.49 (3) & 0.262 (2) & 1.21 (2) \\
    \midrule
    % \rowcolor{EffColour}
     & 1 & 3.78  \lflush{} & 31.50 \lflush{} & 259.71 \lflush{} & 0.101 \lflush{} & 0.44 \lflush{} \\
    % \rowcolor{EffColour}
    Efficient & 4 & 4.71 \lflush{} & 42.16 \lflush{} & 276.15 \lflush{} & 0.170 \lflush{} & 0.72 \lflush{} \\
    % \rowcolor{EffColour}
     & 8 & \textbf{6.10 (1)} & \textbf{58.60 (1)} & \textbf{301.24 (1)} & \textbf{0.256 (1)} & \textbf{1.14 (1)} \\
    \midrule
    % \rowcolor{SupColour}
    & 1 & 4.21 \lflush{} & 33.69 \lflush{} & 264.99 \lflush{} & 0.112 \lflush{} & 0.46 \lflush{} \\
    % \rowcolor{SupColour}
    Super & 4 & 5.07 \lflush{} & 44.47 \lflush{} & 284.49 \lflush{} & 0.178 \lflush{} & 0.74 \lflush{} \\
    % \rowcolor{SupColour}
    & 8 & 6.65 (3) & 60.73 (3) & 309.72 (2) & 0.264 (3) & 1.19 (2) \\
    \bottomrule
  \end{tabular}
  }
\end{table*}
% \end{wraptable}
%

%
%%
\subsection{Speed and Efficiency Comparison}
\label{app: FLOPs}
In the main body and other sections of the Appendix, we present
comprehensive theoretical comparisons and rigorous experiments on
Vision and NLP classification tasks as well as for English-to-Spanish
translation to compare the attention algorithms. Optimized Attention
and Efficient Attention perform on par with standard attention with
25\% and 50\% less parameters respectively. In addition, Super
Attention outperformed all other algorithms significantly while having
25\% fewer parameters compared to standard attention.

As mentioned in the main body, according to the definitions of our
proposed algorithms, Efficient, Optimized, and Super Attention
mechanisms perform 2,1, and 1 fewer matrix multiplication per head
compared to standard attention respectively. Here, we further analyze
and compare the required number of FLOPs for completing a single
forward and backward pass for all algorithms under study to gain
further insight into the efficiency of the proposed algorithms.

\paragraph{FLOPs Versus Projection Dim.} As depicted in \cref{fig:
  FLOPs Versus Proj Dim}, we compare the number of required FLOPs by
each attention algorithm when we fixate the sequence length (denoted
as $\ell$) and vary the projection dimension. Even though the number
of FLOPs scales linearly with the projection dimension for all
algorithms, the slope of this increase differs significantly for each
algorithm. Specifically, for Efficient Attention, the slope of the
line is equal to $9\ell$ while for both Optimized and Super Attention
this is equal to $12 \ell$ compared to $15\ell$ for standard
attention. This means that as we scale the projection dimension the
FLOPs required for finishing a forward and backward pass using
Efficient Attention increases \nicefrac{3}{5} as fast as standard
attention.

\begin{figure}[h!tbp]
  \centering
  \includegraphics[width=0.75\columnwidth]{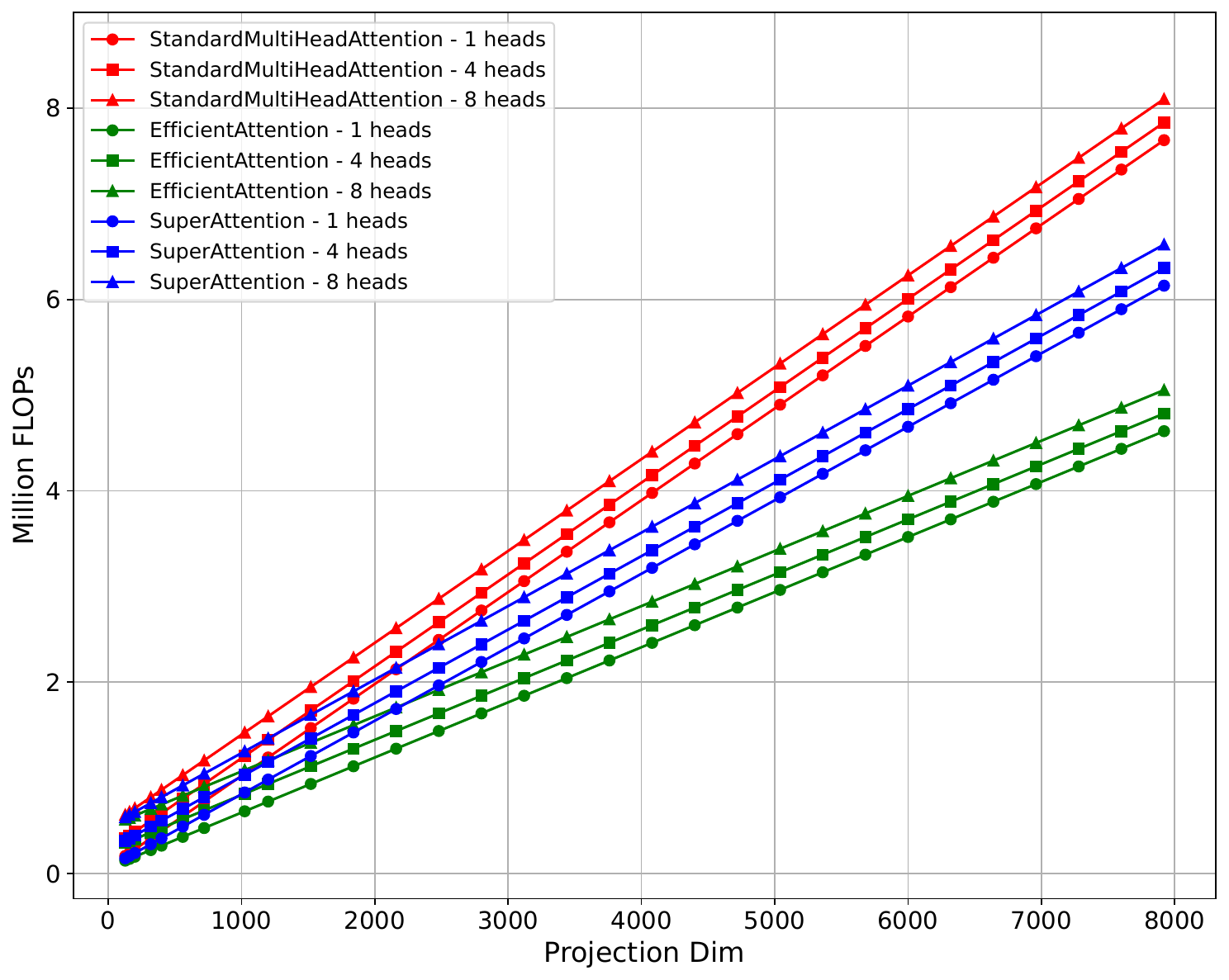}
  \caption{Number of Flops required to complete a single forward plus
    backward pass for each attention mechanism. While the complexity
    and therefore, the number of FLOPs increases linearly as the
    projection dimension increases for all attention mechanisms, the
    slope of the increase varies significantly as depicted in this
    plot. Efficient Attention and Super Attention (Optimized Attention
    is not shown as it is exactly similar to Super Attention) require
    significantly fewer FLOPs as the projection dimension increases
    compared to standard attention. Here sequence length is set to 64
    ($\ell=64$). Trying different values for $\ell$ changes the scale
    of the $y$-axis but the chart looks the same.}
  \label{fig: FLOPs Versus Proj Dim}
\end{figure}
\paragraph{FLOPs Equation.} The number of FLOPs required for finishing
a forward and backward pass for each of the attention mechanisms is
calculated according to the following equation:
\begin{equation}
  \label{eq: FLOPs Formula}
  \begin{split}
    \text{FLOPs} & = C_{\text{Attn}} \ell d_m + 15 h \ell^2
  \end{split}
\end{equation}
where $C_{\text{Attn}}$ is the attention algorithm constant which is
15 for standard attention, 12 for Optimized and Super Attention, and 9
for Efficient Attention, and $\ell$, $d_m$, and $h$ represent the
sequence length, projection dimension, and number of heads consistent
with the notation used throughout the paper.

\cref{fig: Flops All Archs} shows the 3D plot summarizing the number
of FLOPs for each attention algorithm under varying sequence length
and projection dimension in the single head setting. As evident in
\cref{fig: Flops All Archs} and \cref{eq: FLOPs Formula}, our proposed
algorithms need fewer FLOPs as sequence length increases, which is an
important consideration for use in LLMs.

\begin{figure*}[t]
  \centering
  \begin{minipage}[b]{0.45\textwidth}
    \centering
    \includegraphics[width=\textwidth]{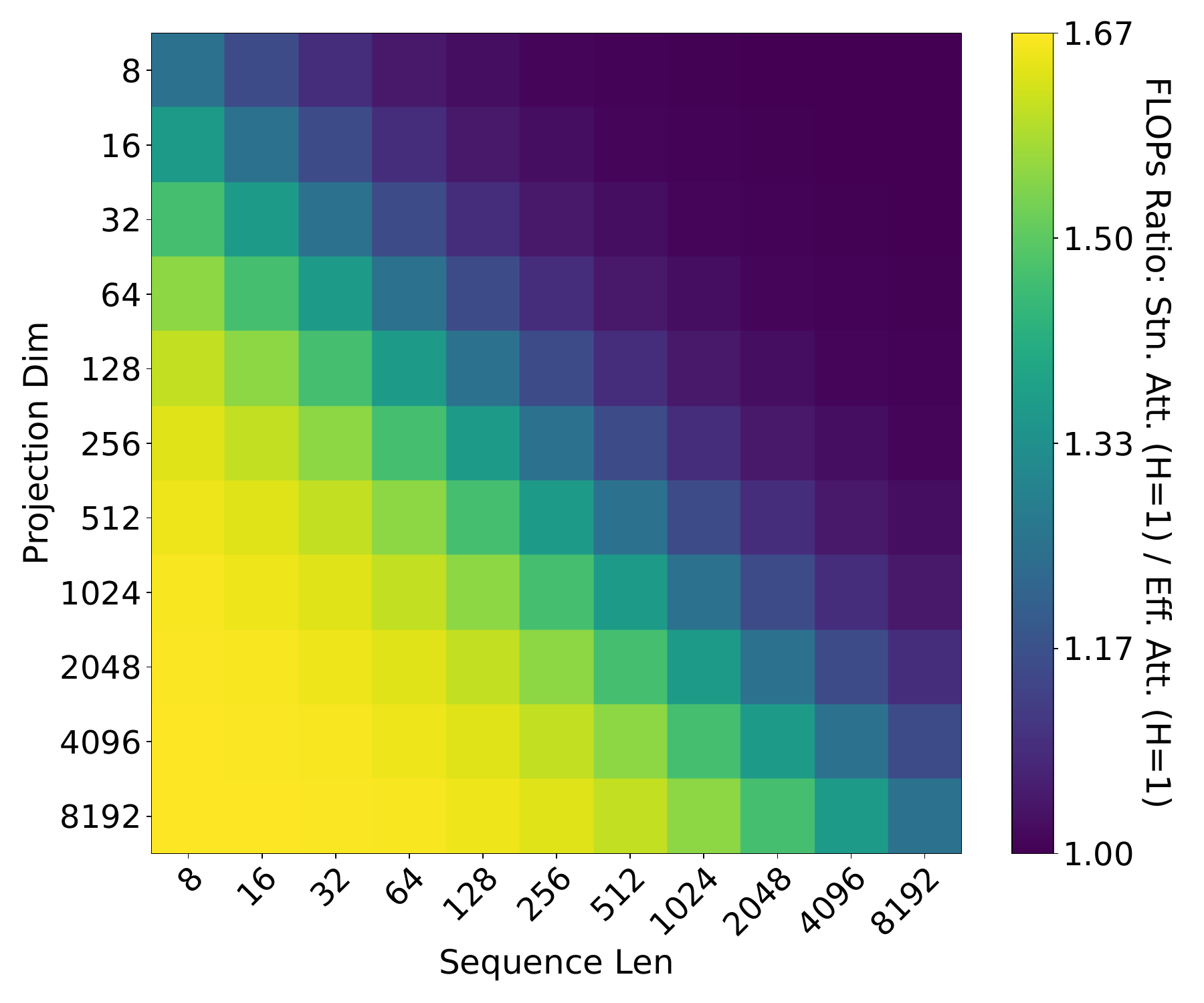}
  \end{minipage}
  \hfill
  \begin{minipage}[b]{0.45\textwidth}
    \centering
    \includegraphics[width=\textwidth]{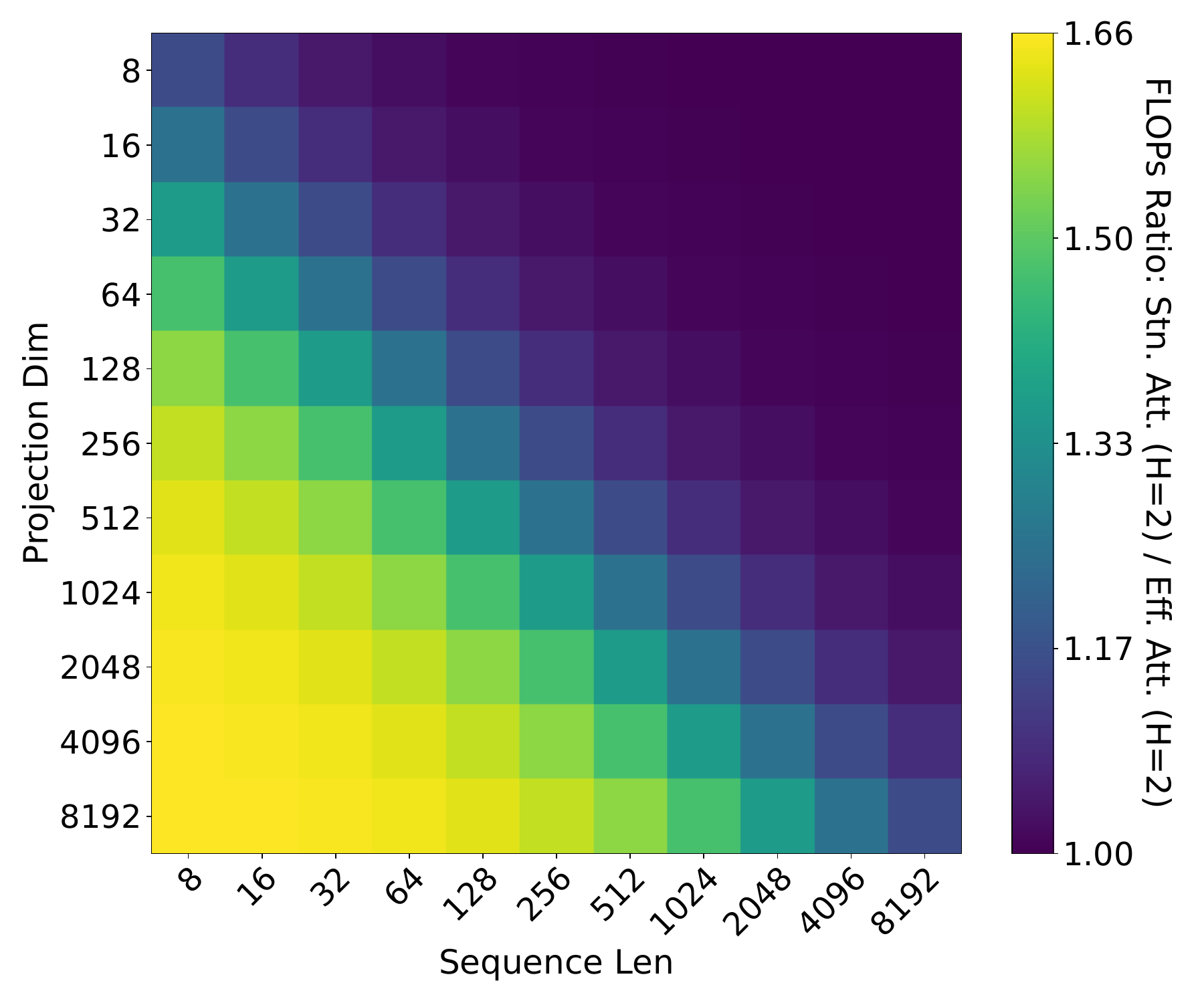}
  \end{minipage}
  \hfill
  \begin{minipage}[b]{0.45\textwidth}
    \centering
    \includegraphics[width=\textwidth]{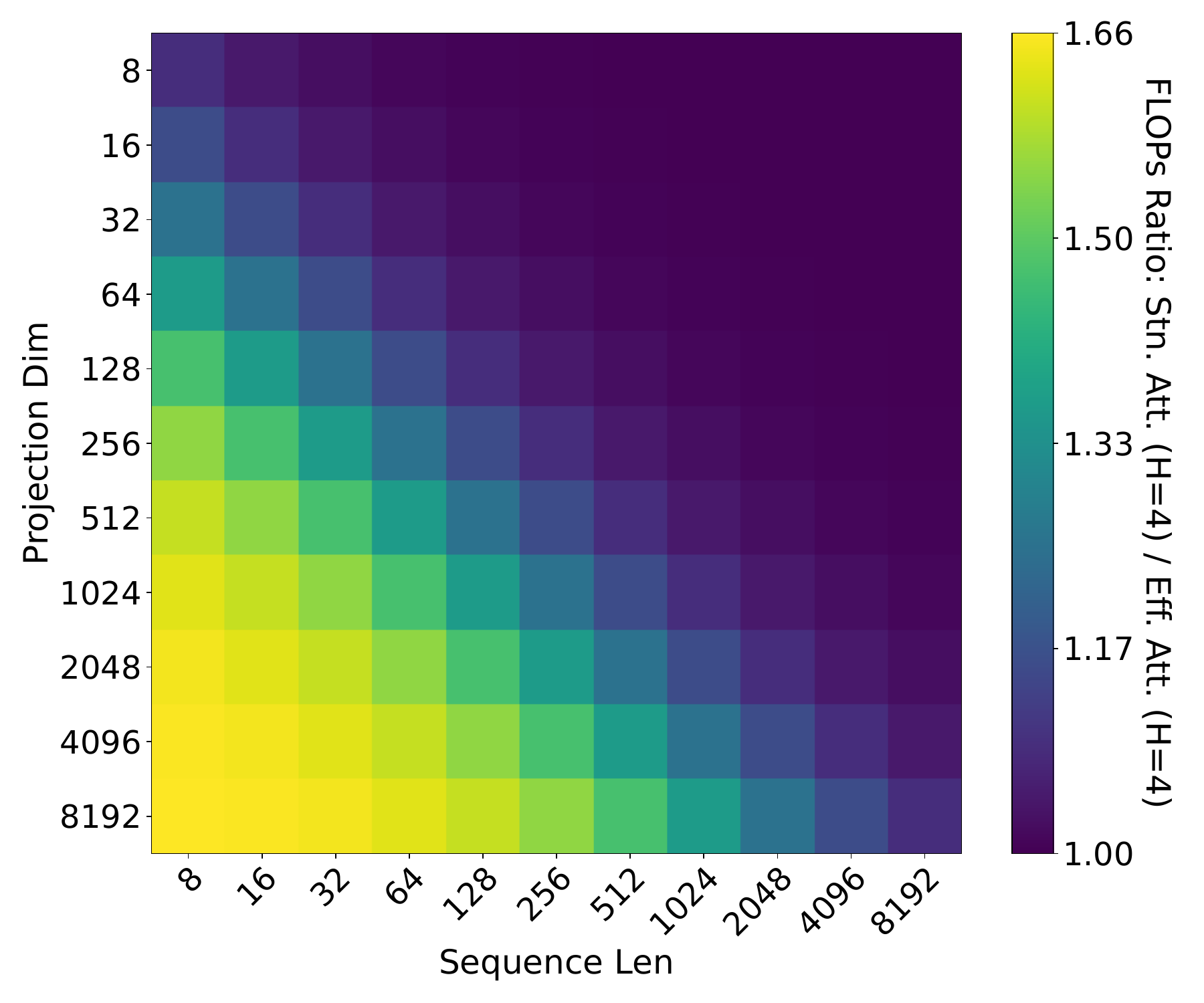}
  \end{minipage}
  \hfill
  \begin{minipage}[b]{0.45\textwidth}
    \centering
    \includegraphics[width=\textwidth]{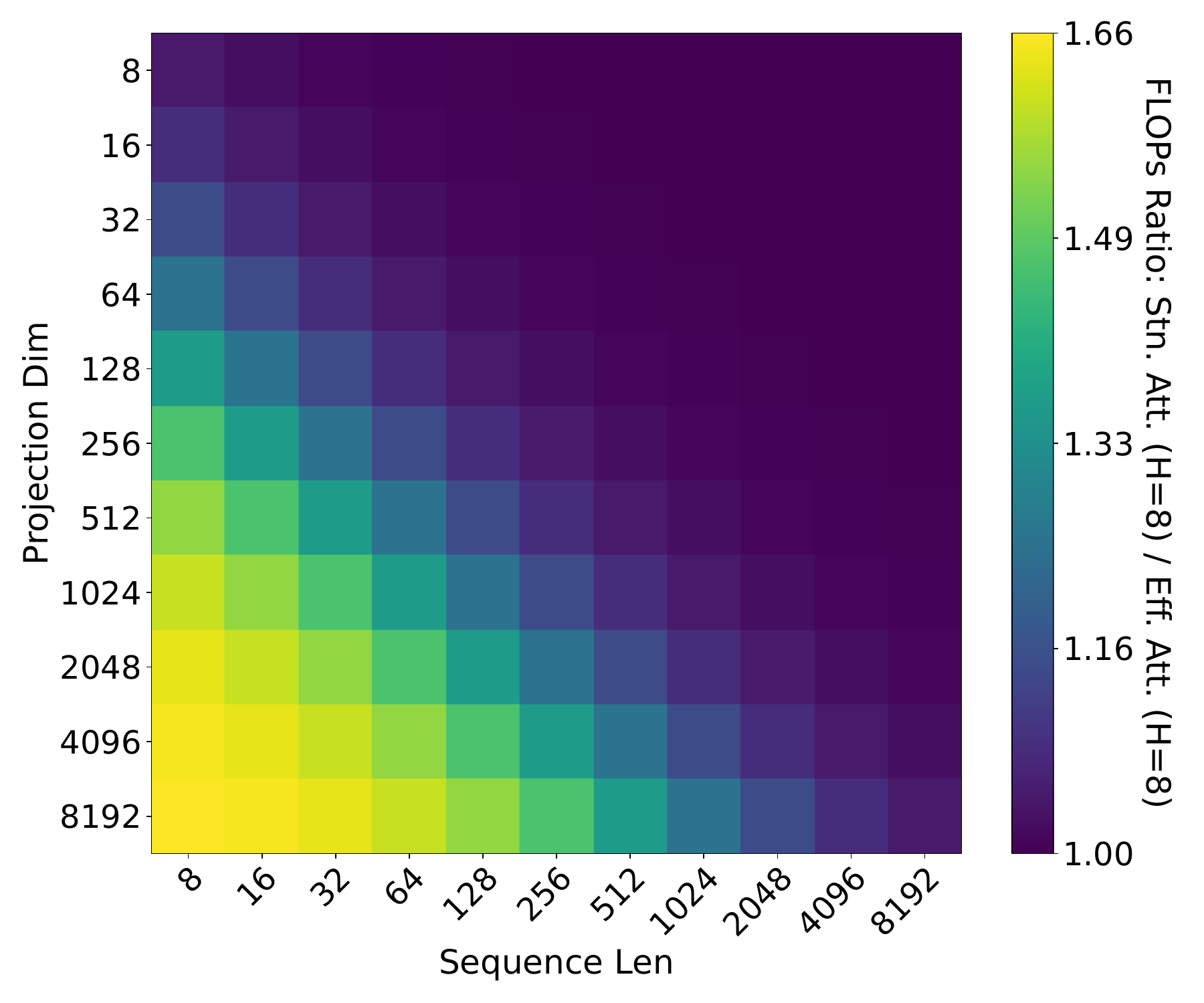}
  \end{minipage}
  \caption{Heatmaps showing the ratio of FLOPs Standard Attention
    requires compared to the Efficient Attention in 1, 2, 4, and 8
    attention head settings. Standard attention requires up to 67\%
    more FLOPs to complete a single forward and backward pass. On
    average, standard attention requires 30\%, 25\%, 20\%, and 16\%
    more FLOPs than Efficient Attention when using 8, 4, 2, and 1
    heads respectively.}
  \label{fig: Flops Heatmaps}
\end{figure*}
\begin{figure*}[th]
  \centering
  \begin{subfigure}[b]{0.24\textwidth}
    \centering
    \includegraphics[width=\textwidth]{flops_single_head/StnAttn_single_head}
    \subcaption{Standard Att.}
  \end{subfigure}
  \hfill
  \begin{subfigure}[b]{0.24\textwidth}
    \centering
    \includegraphics[width=\textwidth]{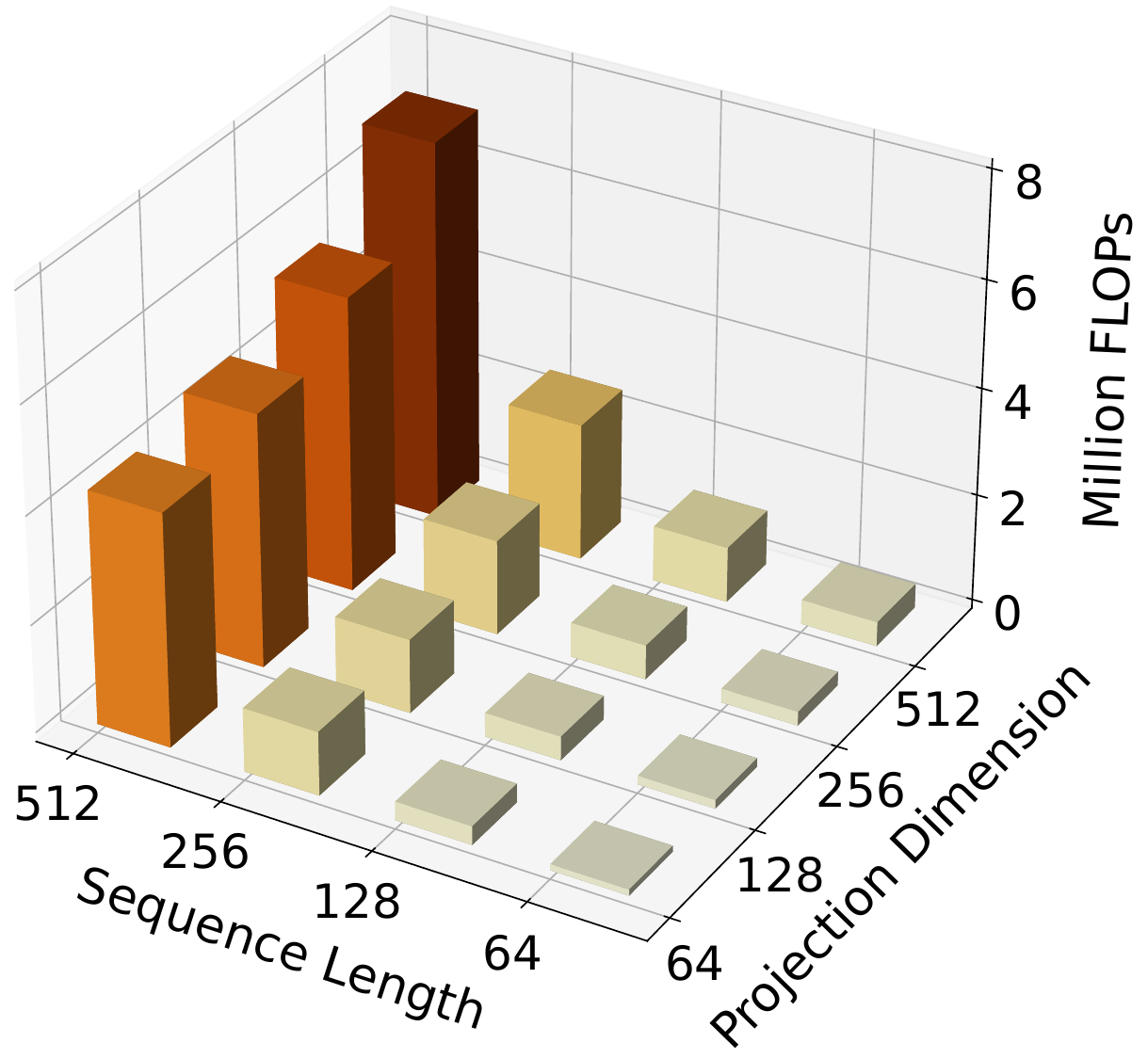}
    \subcaption{Optimized Att.}
  \end{subfigure}
  \hfill
  \begin{subfigure}[b]{0.245\textwidth}
    \centering
    \includegraphics[width=\textwidth]{flops_single_head/EffAttn_single_head}
    \subcaption{Efficient Att.}
  \end{subfigure}
  \hfill
  \begin{subfigure}[b]{0.24\textwidth}
    \centering
    \includegraphics[width=\textwidth]{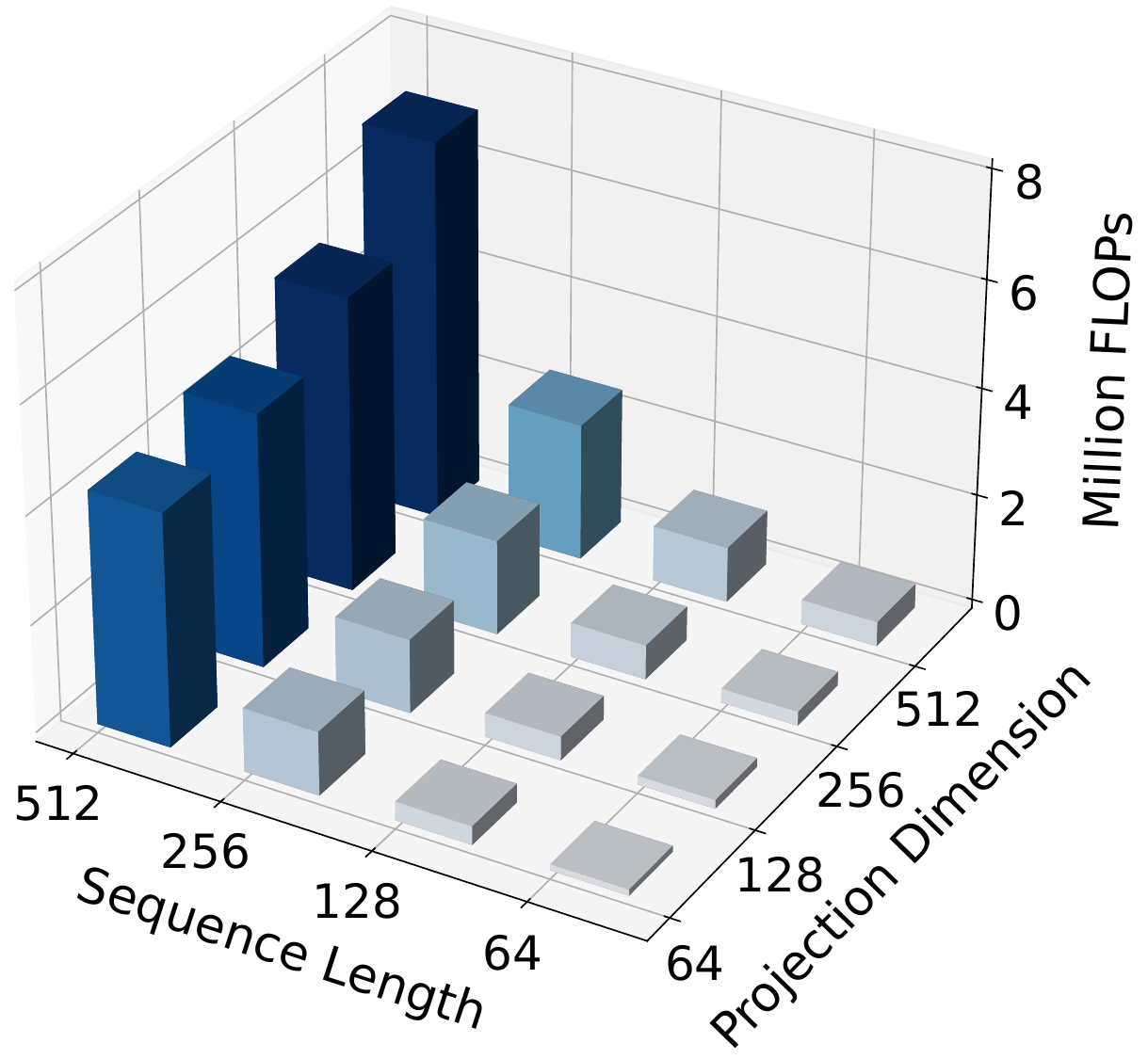}
    \subcaption{Super Att.}
  \end{subfigure}
  \vspace{-0.45em}
  \caption{3D plots visualizing the number of FLOPs for each
    variant in a forward + backward pass given different sequence lengths
    and projection dimensions in single-head setting. Efficient Att. followed by Super and
    Optimized Att. needs substantially fewer FLOPs for completing
    a forward + backward pass compared to standard attention.}
  \label{fig: Flops All Archs Full}
  \vspace{-1.1em}
\end{figure*}
\begin{figure*}[th]
  \centering
  \begin{subfigure}[b]{\textwidth}
    \centering
    \includegraphics[width=0.5\textwidth]{figs/scaling/scaling_laws_legend.pdf}
  \end{subfigure}
  \hfill
  \begin{subfigure}[b]{\textwidth}
    \centering
    \includegraphics[width=\textwidth]{figs/scaling/scaling_laws_test.pdf}
  \end{subfigure}
  \begin{subfigure}[b]{\textwidth}
    \centering
    \includegraphics[width=\textwidth]{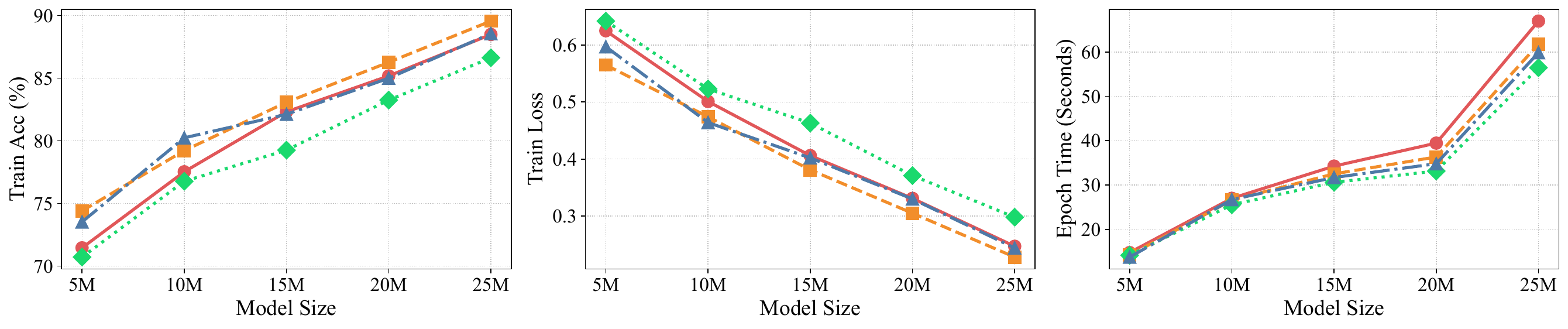}
  \end{subfigure}
  \caption{Performance of different architectures on the Amazon Reviews Classification task as the size of the models increases from 5 Million parameters to 25 Million parameters. The results point to the overparameterization of the Standard Attention as it puts an additional computational burden which is not accompanied with better performance in terms of accuracy or loss.}
  \label{fig: Scaling All}
  % \vspace{0.5em}
\end{figure*}
\paragraph{FLOPs Heatmaps.} In addition to the previous analyses, in
\cref{fig: Flops Heatmaps}, we compare the ratio of FLOPs required to
finish a single forward and backward pass by standard attention to
Efficient Attention under different settings (i.e., varying sequence
length and projection dimension) for different number of heads. In all
scenarios, standard attention requires up to 66\% more FLOPs in
comparison to Efficient Attention. On average, Standard Efficient
requires 30\%, 25\%, 20\%, and 16\% more FLOPs in comparison to
Efficient Attention when using 1, 2, 4, and 8 heads, respectively.

\subsection{Vision Transformers}
\label{app: Vision}

\paragraph{MNIST.} We trained ViT models with different attention
mechanisms, all with two attention layers and model dimension
\(d_m = 128\). As expected, Super Attention outperforms all other
architectures, in terms of accuracy, by at least \(2.68\%\) and
standard attention by \(3.23\%\). The smallest attention layer size
belongs to Efficient Attention, which performs on par with standard
attention. The complete results are presented in \cref{tbl: MNIST}.
\begin{table*}[h]
  \centering
  \caption{Averages of different metrics over five runs in the MNIST
    experiment. The numbers in parentheses indicate the ranking of
    each mechanism for that metric. An ablation study on the number of
    heads shows increasing the number of heads enhances the
    performance of all algorithms. As expected, the Efficient
    Attention model has the smallest attention layer size and the
    Super Attention model performs the best in terms of accuracy and
    loss.}
  \label{tbl: MNIST}
  \smallskip
  \resizebox{\textwidth}{!}{%
  \begin{tabular}{lrrrrccccc}
    \toprule
    Att. & $h$ & $d_m$ & $d_k$ & \# Param. & Avg. Time (s) & Acc. (\%) & Loss & Val Acc. (\%) & Val Loss\\
    \midrule
    % \rowcolor{StnColour}
          & 1 & 128 & 128 & 66,048 \lflush{} & 8.15 \lflush{} & 93.26  \lflush{} & 0.227 \lflush{} & 98.02 \lflush{} & 0.063 \lflush{} \\
    % \rowcolor{StnColour}
    Stn.  & 2 & 128 & 64 & 66,048 \lflush{} & 8.18 \lflush{} & 95.40 \lflush{} & 0.161 \lflush{} & 98.61 \lflush{} & 0.049 \lflush{} \\
    % \rowcolor{StnColour}
          & 4 & 128 & 32 & 66,048 (4) & 8.31 (4) & 93.73 (4) & 0.209 (4) & 98.12 (4) & 0.062 (4) \\
    \midrule
    % \rowcolor{OptColour}
          & 1 & 128 & 128 & 49,536 \lflush{} & 7.56 \lflush{} & 91.02 \lflush{} & 0.299 \lflush{} & 97.30 \lflush{} & 0.095 \lflush{} \\
    % \rowcolor{OptColour}
    Opt.  & 2 & 128 & 64 & 49,536 \lflush{} & 7.57 \lflush{} & 93.70 \lflush{} & 0.215 \lflush{} & 97.93 \lflush{} & 0.071 \lflush{} \\
    % \rowcolor{OptColour}
      & 4 & 128 & 32 & 49,536 (3) & 7.68 (3) & 95.36 (2) & 0.161 (2) & 98.43 (2) & 0.046 (2)\\
    \midrule
    % \rowcolor{EffColour}
    & 1 & 128 & 128 & 33,024 \lflush{} & 6.89 \lflush{} & 93.29 \lflush{} & 0.228 \lflush{} & 97.78 \lflush{} & 0.073 \lflush{} \\
    % \rowcolor{EffColour}
    Eff. & 2 & 128 & 64 & 33,024 \lflush{} & 6.99 \lflush{} & 93.60 \lflush{} & 0.223 \lflush{} & 98.11 \lflush{} & 0.061 \lflush{} \\
    % \rowcolor{EffColour}
    & 4 & 128 & 32 & \textbf{33,024 (1)}  & \textbf{7.05 (1)} & 94.28 (3) & 0.197 (3) & 98.27 (3) & 0.058 (3) \\
    \midrule
    % \rowcolor{SupColour}
    & 1 & 128 & 128 & 37,184 \lflush{} & 7.46 \lflush{} & 96.24 \lflush{} & 0.136 \lflush{} & 98.32 \lflush{} & 0.056 \lflush{} \\
    % \rowcolor{SupColour}
    Sup.  & 2 & 128 & 64 & 37,184 \lflush{} & 7.50 \lflush{} & 96.59 \lflush{} & 0.124 \lflush{} & 98.52 \lflush{} & 0.050 \lflush{} \\
    % \rowcolor{SupColour}
    & 4 & 128 & 32 & 37,184 (2) & 7.58 (2) & \textbf{96.96 (1)} & \textbf{0.112 (1)} & \textbf{98.62 (1)} & \textbf{0.051 (1)} \\
    \bottomrule
  \end{tabular}%
  }
\end{table*}
\paragraph{ImageNet.} Scaling the vision experiments even further, the
ImageNet1k dataset presents much more complexity as the labels
comprise 1000 classes. We used a modified ViT-B/16 model architecture,
employed different attention mechanisms in its Transformers blocks,
and trained the models. Due to our computational constraints, we
reduced the number of transformer blocks from 12 to 8, resized the
images to $112\!\times\!112$ (instead of the original
$224\!\times\!224$) and reduced the patch size from 16 to 8 to enable
training on our Nvidia RTX 4090 GPU. Other parameters are similar to
the original architecture; specifically, \(d_m=768\) and
\(h=12\). \cref{tbl: Vision,tbl: ImageNet} present the results of our
experiments on the ImageNet dataset. 
% In both experiments, Super
% Attention significantly outperforms standard attention (boosting train
% accuracy 27\% and 17.25\% in \cref{tbl: Vision,tbl: ImageNet}
% experiments respectively).
%
\begin{table*}[th]
  \centering
  \caption{Performance of different architectures on the ImageNet
    dataset. Since different attention layer architectures in the main
    ImageNet experiment had different numbers of parameters, an
    interesting ablation study is comparing these architectures when
    the total number of parameters is very close. To achieve this, we
    change some hyperparameters like $d_m$ or the number of attention
    layers from the previous experiment. The numbers in parentheses
    indicate the ranking of each mechanism for that metric. We used a
    modified ViT-B/16 model, plugged in the attention algorithms in
    the Transformers block, and trained the models. Super Attention
    significantly outperforms all other algorithms. Unlike the results reported in \cref{tbl: Vision} in the main body, the models in this ablation experiment are not pre-trained on ImageNet21K (as such the accuracies and validation accuracies are lower compared to the ones with pre-training).}
  \label{tbl: ImageNet}
  \smallskip
  \resizebox{\textwidth}{!}{%
  \begin{tabular}{lrrcrcccccc}
    \toprule
    Att. & $h$ & $d_m$ & Att. Layers & Tot. \# Param. & Acc. (\%) & Loss & Top 5 & Val Acc. (\%) & Val Loss & Val Top 5\\
    \midrule
    % \rowcolor{StnColour}
         Stn. & 12 & 768 & 8 & 60.54M (4) & 51.18 (4) &  2.09 (4) & 76.05 (4) & 32.74 (4) & 3.36 (4) & 56.48 (4) \\
    \midrule
    % \rowcolor{OptColour}
        Opt.  & 12 & 816 & 8 & 60.12M (2) & 53.22 (2) & 1.98 (2) & 77.21 (2) & 33.44 (3) & 3.23 (3) &  57.37 (3) \\
    \midrule
    % \rowcolor{EffColour}
    Eff. & 12 & 804 & 9  & \textbf{60.09M (1)} & 51.28 (3) & 2.06 (3) & 76.66 (3) & \textbf{35.49 (1)} & \textbf{3.13 (1)} & \textbf{59.69 (1)} \\
    \midrule
    % \rowcolor{SupColour}
    Sup. & 12 & 804 & 9 & 60.44M (3) & \textbf{64.98 (1)} & \textbf{1.37 (1)} & \textbf{87.36 (1)} & 34.31 (2) & 3.18 (2) & 58.70 (2) \\
    \bottomrule
  \end{tabular}%
  }
\end{table*}

Val. results in \cref{tbl: Vision,tbl: MNIST,tbl: ImageNet} refer to
models' performances  on the official validation set for ImageNet1K,
and the official tests sets for MNIST and CIFAR100 datasets.

\subsection{Natural Language Processing}
\label{app: NLP}

\subsubsection{Transformer for Text Classification}

\paragraph{IMDB.} The IMDB dataset includes 50,000 reviews with binary
labels, indicating negative and positive sentiments. The Transformer
models, used in this experiment, all have a single attention layer
with model dimension and context length 32. The complete results are
presented in \cref{tbl: IMDB}.

\begin{table*}[th]
  \centering
  \caption{Averages of different metrics over five runs in the IMDB
    experiment. Here, varying the number of heads doesn't meaningfully
    affect the performance of any of the algorithms. As expected, the
    Efficient Attention model has the smallest attention layer size
    and the Super Attention model performs the best in terms of
    accuracy and loss.}
  \label{tbl: IMDB}
  \smallskip
  \resizebox{\textwidth}{!}{%
  \begin{tabular}{lrrrcccccc}
    \toprule
    Att. & $h$ & $d_m$ & $d_k$ & \# Param. & Avg. Time & Acc. (\%) & Loss & Test Acc. (\%) & Test Loss\\
    \midrule
    % \rowcolor{StnColour}
         & 1 & 32 & 32 & 4,224 \lflush{} & 0.284 \lflush{} & 96.09 \lflush{} & 0.082 \lflush{} & 78.09 \lflush{} & 0.461 \lflush{} \\
    % \rowcolor{StnColour}
    Stn. & 2 & 32 & 16 & 4,224 \lflush{} & 0.297 \lflush{} & 95.51 \lflush{} & 0.112 \lflush{} & 78.14 \lflush{} & 0.467 \lflush{} \\
    % \rowcolor{StnColour}
         & 4 & 32 & 8 & 4,224 (4) & 0.315 (4) & 95.70 (4) & 0.086 (3) & 77.62 (4) & 0.474 (4) \\
    \midrule
    % \rowcolor{OptColour}
         & 1 & 32 & 32 & 3,168 \lflush{} & 0.283 \lflush{} & 96.62 \lflush{} & 0.070 \lflush{} & 78.00 \lflush{} & 0.461 \lflush{} \\
    % \rowcolor{OptColour}
    Opt. & 2 & 32 & 16 & 3,168 \lflush{} & 0.299 \lflush{} & 96.77 \lflush{} & 0.073 \lflush{} & 78.00 \lflush{} & 0.460 \lflush{} \\
    % \rowcolor{OptColour}
         & 4 & 32 & 8 & 3,168 (2) & 0.305 (3) & 96.31 (3) & 0.095 (4) & 77.85 (2) & 0.472 (2) \\
    \midrule
    % \rowcolor{EffColour}
    & 1 & 32 & 32 & 2,112 \lflush{} & 0.267 \lflush{} & 96.66 \lflush{} & 0.080 \lflush{} & 77.58 \lflush{} & 0.478 \lflush{} \\
    % \rowcolor{EffColour}
    Eff. & 2 & 32 & 16 & 2,112 \lflush{} & 0.273 \lflush{} & 96.86 \lflush{} & 0.068 \lflush{} & 77.74 \lflush{} & 0.473 \lflush{} \\
    % \rowcolor{EffColour}
    & 4 & 32 & 8 & \textbf{2,112 (1)} & \textbf{0.280 (1)} & 96.41 (2) & \textbf{0.064 (1)} & 77.77 (3) & \textbf{0.468 (1)} \\
    \midrule
    % \rowcolor{SupColour}
    & 1 & 32 & 32 & 3,168  \lflush{} & 0.272  \lflush{} & 97.68  \lflush{} & 0.063  \lflush{} & 78.21  \lflush{} & 0.472  \lflush{} \\
    % \rowcolor{SupColour}
    Sup. & 2 & 32 & 16 & 3,168  \lflush{} & 0.294 \lflush{} & 97.84 \lflush{} & 0.064 \lflush{} & 78.35 \lflush{} & 0.454 \lflush{} \\
    % \rowcolor{SupColour}
    & 4 & 32 & 8 & 3,168 (2) & 0.299 (2) & \textbf{97.45 (1)} & 0.070 (2) & \textbf{78.34 (1)} & 0.472 (2)\\
    \bottomrule
  \end{tabular}%
  }
\end{table*}
\paragraph{Amazon Reviews.} The Amazon Reviews dataset poses a
different challenge than the IMDB dataset as it is a significantly
larger dataset with 3,650,000 reviews, containing a wider range of
sentiments in \(1, 2, \dots, 5\); higher values indicate more positive
sentiment. The Transformer models, used in this experiment, all have
three attention layers with model dimension and context length 64. The
complete results are presented in \cref{tbl: Amazon}.

\begin{table*}[h]
  \centering
  \caption{Averages of different metrics over five runs in the Amazon
    Reviews experiment. An ablation study on the number of heads shows
    increasing the number of heads helps improve the performance of
    all algorithms. The Efficient Attention model has the smallest
    attention layer size and the Super Attention model performs the
    best in accuracy and loss.}
  \label{tbl: Amazon}
  \smallskip
  \resizebox{\textwidth}{!}{%
  \begin{tabular}{lrrrrccccc}
    \toprule
    Att. & $h$ & $d_m$ & $d_k$ & \# Param. & Avg. Time & Acc. & Loss & Val Acc. & Val Loss\\
    \midrule
    % \rowcolor{StnColour}
         & 1 & 128 & 128 & 66,048 \lflush{} & 42.06 \lflush{} & 86.76 \lflush{} & 0.31 \lflush{} & 62.32 \lflush{} & 0.87 \lflush{} \\
    % \rowcolor{StnColour}
    Stn. & 2 & 128 & 64 & 66,048 \lflush{} & 50.91 \lflush{} & 87.13 \lflush{} & 0.30 \lflush{} & 63.66 \lflush{} & 0.81 \lflush{} \\
    % \rowcolor{StnColour}
         & 4 & 128 & 32 & 66,048 (4) & 66.97 (4) & 88.49 (3) & 0.25 (3) & 65.55 (4) & 0.77 (3)\\
    \midrule
    % \rowcolor{OptColour}
         & 1 & 128 & 128 & 49,536 \lflush{} & 38.68 \lflush{} & 89.41 \lflush{} & 0.25 \lflush{} & 63.03 \lflush{} & 0.82 \lflush{} \\
    % \rowcolor{OptColour}
    Opt. & 2 & 128 & 64 & 49,536 \lflush{} & 47.97 \lflush{} & 90.48 \lflush{} & 0.23 \lflush{} & 65.98 \lflush{} & 0.75 \lflush{} \\
    % \rowcolor{OptColour}
         & 4 & 128 & 32 & 49,536 (3) & 61.75 (3) & \textbf{89.56 (1)} & \textbf{0.23 (1)} & 65.67 (2) & 0.75 (2) \\
    \midrule
    % \rowcolor{EffColour}
    & 1 & 128 & 128 & 33,024 \lflush{} & 34.82 \lflush{}  & 88.56 \lflush{} & 0.27 \lflush{} & 63.42 \lflush{} & 0.81 \lflush{} \\
    % \rowcolor{EffColour}
    Eff. & 2 & 128 & 64 & 33,024 \lflush{} & 42.19 \lflush{} &  88.36 \lflush{} & 0.27 \lflush{} & 63.81 \lflush{} & 0.80 \lflush{} \\
    % \rowcolor{EffColour}
     & 4 & 128 & 32 & \textbf{33,024 (1)} & \textbf{56.44 (1)} & 86.63 (4) &  0.29 (4) & 65.58 (3) & 0.77 (3) \\
    \midrule
    % \rowcolor{SupColour}
    & 1 & 128 & 128 & 42,336 \lflush{} & 37.78 \lflush{} & 89.11 \lflush{} & 0.26 \lflush{} & 65.73 \lflush{} & 0.74 \lflush{} \\
    % \rowcolor{SupColour}
    Sup.  & 2 & 128 & 64 & 42,336 \lflush{} & 46.24 \lflush{} & 88.41 \lflush{} & 0.27 \lflush{} & 67.22 \lflush{} & 0.73 \lflush{} \\
    % \rowcolor{SupColour}
    & 4 & 128 & 32 & 42,336 (2) & 59.86 (2) & 88.56 (2) & 0.24 (2)  & \textbf{68.10 (1)} & \textbf{0.71 (1)} \\
    \bottomrule
  \end{tabular}%
  }
\end{table*}
\subsubsection{Transformer for Machine Translation}

\begin{table*}[h!]
  \centering
  \caption{Averages of different metrics over five runs trained on
    Europarl and Anki English-to-Spanish translation datasets. The
    numbers in parentheses indicate the ranking of each mechanism for
    that metric. An ablation study on the number of heads shows
    increasing the number of heads enhances the performance of all
    algorithms. Optimized and Efficient Attentions perform on par or
    better than Standard Attention on most benchmarks with 1/2 and 3/4
    as many attention parameters.}
  \label{tbl: Translation Complete}
  \smallskip
  \resizebox{\textwidth}{!}{%
  \begin{tabular}{lrrrrccccccc}
    \toprule
    Att. & $h$ & $d_m$ & $d_k$ & \# Param. & Avg. Time & BLEU & Acc. & Loss & Val BLEU & Val Acc. & Val Loss\\
    \midrule
    % \rowcolor{StnColour}
         & 1 & 1024 & 1024 & 4,198,400 \lflush{} & 556.5 \lflush{} & 23.2 \lflush{} & 80.48 \lflush{} & 0.86 \lflush{} & 22.1 \lflush{} & 80.86 \lflush{} & 0.87\lflush{} \\
    % \rowcolor{StnColour}
    Stn. & 2 & 1024 & 512 & 4,198,400 \lflush{} & 598.7 \lflush{} & 22.3 \lflush{} & 81.03\lflush{} & 0.84 \lflush{} & 22.7 \lflush{} & 81.43 \lflush{} & 0.84 \lflush{} \\
    % \rowcolor{StnColour}
         & 4 & 1024 & 256 & 4,198,400 (3) & 600.0 (3) & 23.1 (2) & 81.11 (3) & 0.83 (3) & \textbf{22.8 (1)} & 81.41 (3) & 0.84 (3) \\
    \midrule
    % \rowcolor{OptColour}
         & 1 & 1024 & 1024 & 3,148,800 \lflush{} & 552.0 \lflush{} & 22.5 \lflush{} & 81.15 \lflush{} & 0.87 \lflush{} & 22.6 \lflush{} & 81.11 \lflush{} & 0.84\lflush{} \\
    % \rowcolor{OptColour}
    Opt. & 2 & 1024 & 512 & 3,148,800 \lflush{} & 583.8 \lflush{} & 22.1 \lflush{} & 81.61 \lflush{} & 0.82 \lflush{} & 23.0 \lflush{} & 81.57 \lflush{} & 0.82 \lflush{} \\
    % \rowcolor{OptColour}
         & 4 & 1024 & 256 & 3,148,800 (2) & 586.8 (2) & \textbf{24.5 (1)} & \textbf{82.06 (1)} & \textbf{0.78 (1)} & 22.6 (3) & \textbf{81.98 (1)} & \textbf{0.80 (1)} \\
    \midrule
    % \rowcolor{EffColour}
    & 1 & 1024 & 1024 & 2,099,200 \lflush{} & 472.7 \lflush{} & 22.4 \lflush{} & 81.13 \lflush{} & 0.82 \lflush{} & 22.8 \lflush{} & 81.43 \lflush{} & 0.83 \lflush{} \\
    % \rowcolor{EffColour}
    Eff. & 2 & 1024 & 512 & 2,099,200 \lflush{} & 498.6 \lflush{} & 22.3 \lflush{} & 81.48 \lflush{} & 0.80 \lflush{} & 22.9 \lflush{} & 81.62 \lflush{} & 0.81 \lflush{} \\
    % \rowcolor{EffColour}
    & 4 & 1024 & 256 & \textbf{2,099,200 (1)} & \textbf{523.0 (1)}  & 22.6 (3) & 81.15 (2) & 0.82 (2) & 22.3 (3) & 81.44 (2) & 0.83 (2) \\
    \bottomrule
  \end{tabular}%
  }
\end{table*}
\paragraph{Europarl Parallel Corpus and Anki.} Anki dataset for
English-Spanish translation consists of more than 118,000 sentence
pairs in both English and Spanish languages. While training a model on
this dataset enables basic translation, the educational nature and
size of the dataset are too simple for training a capable translation
model. Therefore, we also add the Europarl Parallel Corpus which has
around 2 million examples in both English and Spanish languages and
has sentences with much more technical and sophisticated terms to
enable training in a powerful English-to-Spanish translation model. We
then shuffle the mix of both datasets, and randomly split the dataset
into 99.8\%, 0.1\%, and 0.1\% for train, validation, and test splits
respectively.

We then train a translation model inspired by the implementation
available on the official Keras website for translation but with 2
decoder blocks and one encoder block for 6 epochs. Additionally, we
set the \(d_m=1024\) and try 1, 2, and 4 as the number of heads. We
use Sparse Categorical Cross Entropy as our loss metric. The complete
analysis of the results is available in \cref{tbl: Translation
  Complete}.

All 3 algorithms perform comparably in terms of BLEU score, Accuracy,
and Loss. However, the number of attention parameters per
encoder/decoder layer is \nicefrac{1}{2} and \nicefrac{3}{4} of
standard attention in Efficient and Optimized Attention
respectively. Additionally, Efficient attention is up to
\(\nicefrac{(556.5 - 472.7)}{556.6} =15.06\%\) faster to train in
comparison to the standard attention.

\subsection{Evaluation For Use in LLMs}
\label{subsec: LLM}
In addition to evaluating the standard SDPA and its variants for generative language modelling in a scale of around ~125M parameters, we also trained a Language Model (LM) with 1.1B parameters based on Efficient Attention architecture to see the feasibility and scalability of this variant of SDPA in a large scale experiment. This Language Model achieves lower loss than the similarly-sized TinyLlama model, which is based on Standard Attention (details are provided in \cref{table: Large LM} below). We could not train more LMs based on other architectures due to our limited computational resources. The LM based on Efficient Attention was trained using a GPU credit donation that we used to train our LM over 8 weeks on 30 billion tokens of C4 dataset \citep{Raffe+19} using a single A100 with 80GB of GPU.

\begin{table*}[th]
  \centering
  \caption{A Language Model (Based on Efficient Attention) compared to TinyLlama (Based on Standard Attention) after training on 30 billion tokens of C4 dataset. We set the number of heads to 1 in this LM to make training faster. Despite this, this LM performs favourably (5.8$\%$ smaller categorical cross-entropy loss) compared to TinyLlama.}
  \label{table: Large LM}
  \smallskip
  \resizebox{0.85\textwidth}{!}{%
  \begin{tabular}{lccccc}
    \toprule
    name & \# layers & \# heads & model dim & intermediate size & loss \\
    \midrule
    % \rowcolor{StnColour}
        TinyLlama & 22 & 32 & 2048 & 5632 & 2.25 \\
    % \rowcolor{EffColour}
        Efficient based LM & 10 & 1 & 3072 & 8192 & 2.12 \\
    \bottomrule
  \end{tabular}%
  }
\end{table*}

\section{Additional Related Work}
\label{app: Related Work 2}
Flash Attention \citep{Dao+22} and Flash Attention 2 \citep{Dao23}
optimize multi-head attention for modern GPUs without changing its
structure, enabling faster processing and reduced memory demands. It's
worth mentioning our proposed algorithms also benefit from these
optimizations.

With the adoption of LLMs and Foundation Models (FMs), a lot of work
has been done to improve their scalability and deployability. LoRA
\citep{Hu+22} adapts pre-trained models with minimal additional
parameters, and QLoRA \citep{DettmersPHZ23} incorporates quantization
to reduce memory and computational demands.

Quantization has revolutionized the adoption of FMs, particularly
those based on Transformers. Recent advances include mixed-precision
post-training quantization for vision transformers \citep{Liu+21},
quantization-aware training \citep{Jacob+18, Nagel+22},
mixed-precision training \citep{Micikevicius+18}, dynamic quantization
\citep{Zhang+21quantization}, and layer-wise quantization
\citep{ChenWP19}.

Moreover, \citet{Ding+22} unveiled a cutting-edge
framework enhancing quantized model accuracy without significant
performance degradation. However, quantization faces challenges such
as potential performance drops and increased vulnerability to
adversarial attacks \citep{HongPKD21, GuptaA22}.

%%% Local Variables:
%%% mode: latex
%%% TeX-master: "../main"
%%% End:

%%%%%%%%%%%%%%%%%%%%%%%%%%%%%%%%%%%%%%%%%%%%%%%%%%%%%%%%%%%%%%%%%%%%%%%%%%%%%%%
%%%%%%%%%%%%%%%%%%%%%%%%%%%%%%%%%%%%%%%%%%%%%%%%%%%%%%%%%%%%%%%%%%%%%%%%%%%%%%%

\end{document}

\begin{tikzpicture}[node distance=5em]
% Main Attention Node
\node (start3) [attention, xshift=0.6em, yshift=0.6em, opacity=0.3] {};
\node (start2) [attention, xshift=0.3em, yshift=0.3em, opacity=0.6] {};
\node (start1) [attention] {Softmax of \\Scaled Dot-Product};

% ********** Below Main ********** %
% ******************************** %
% K Processing
\node (fn-k3) [slice, below of=start1, xshift=3.3em, yshift=0.6em, fill opacity=0.3] {};
\node (fn-k2) [slice, below of=start1, xshift=3.0em, yshift=0.3em, fill opacity=0.6] {};
\node (fn-k1) [slice, below of=start1, xshift=2.7em] {\ours{Slice}};
% V processing
\node (fn-v1) [linstar, right of=fn-k1] {\ours{Linear*}};
\node (fn2-v3) [slice, right of=start3, xshift=2.7em, fill opacity=0.3] {};
\node (fn2-v2) [slice, right of=start2, xshift=2.7em, fill opacity=0.6] {};
\node (fn2-v1) [slice, right of=start1, xshift=2.7em] {\ours{Slice}};
% Q processing
\node (fn-q3) [linear, left of=fn-k1, xshift=0.6em, yshift=0.6em, fill opacity=0.3] {};
\node (fn-q2) [linear, left of=fn-k1, xshift=0.3em, yshift=0.3em, fill opacity=0.6] {};
\node (fn-q1) [linear, left of=fn-k1] {Linear};
% ******************************** %
% ******************************** %

% ********** Above Main ********** %
% ******************************** %
\node (a-dot-v3) [dotprod, above of=start3, opacity=0.3] {};
\node (a-dot-v2) [dotprod, above of=start2, opacity=0.6] {};
\node (a-dot-v1) [dotprod, above of=start1] {Dot Product};
\node (concat) [concat, above of=a-dot-v1, yshift=-0.4em] {Concat};
\node (fn-o) [linear, above of=concat, yshift=-1em] {Linear};
\node (out) [inpout, above of=fn-o, yshift=-1em] {$O$};
% ******************************** %
% ******************************** %

% ****** Arrows Below Main ******* %
% ******************************** %
% QKV
\node (k) [inpout, below of=fn-k1, yshift=1.0em] {$K$};
\node (v) [inpout, below of=fn-v1, yshift=1.0em] {$V$};
\node (q) [inpout, below of=fn-q1, yshift=1.0em] {$Q$};

% QKV processing -> Attention
% K -> Attention
\draw [arrow] (fn-k1) -- (start1.south -| fn-k1.north);
\draw [thick, opacity=0.6] (fn-k2) -- (start1.south -| fn-k2.north);
\draw [thick, opacity=0.3] (fn-k3) -- (start1.south -| fn-k3.north);
% Q -> Attention
\draw [arrow] (fn-q1.north) -- (start1.south -| fn-q1.north);
\draw [thick, opacity=0.6] (fn-q2) -- (start1.south -| fn-q2.north);
\draw [thick, opacity=0.3] (fn-q3) -- (start1.south -| fn-q3.north);
% V -> Dot Product
\draw [arrow, rounded corners=0.5em] (fn-v1.north) -- (fn2-v1.south);
\draw [arrow, rounded corners=0.5em] (fn2-v1.north) |- (a-dot-v1.east);
\draw [arrow, rounded corners=0.5em, opacity=0.6] (fn2-v2.north) |- (a-dot-v2.east);
\draw [arrow, rounded corners=0.5em, opacity=0.3] (fn2-v3.north) |- (a-dot-v3.east);
% Inpus
\draw [arrow] (k) -- (fn-k1);
\draw [arrow] (v) -- (fn-v1);
\draw [arrow] (q) -- (fn-q1);
% ******************************** %
% ******************************** %

% ****** Arrows Above Main ******* %
% ******************************** %

% A -> A.V
\draw [arrow]  (start1.north) -- (a-dot-v1);
\draw [thick, opacity=0.6] (start2.north) -- (a-dot-v1.south -| a-dot-v2.south);
\draw [thick, opacity=0.3] (start3.north) -- (a-dot-v1.south -| a-dot-v3.south);

% A.V -> Concat
\draw [arrow] ([xshift=-0.4em]a-dot-v1.north) -- ([xshift=-0.4em]concat.south -| a-dot-v1.north);
\draw [arrow, opacity=0.6] ([xshift=-0.2em]a-dot-v2.north) -- ([xshift=-0.2em]concat.south -| a-dot-v2.north);
\draw [arrow, opacity=0.3] ([xshift=0.0em]a-dot-v3.north) -- ([xshift=0.0em]concat.south -| a-dot-v3.north);

% Concat -> Output Linear
\draw [arrow] (concat) -- (fn-o);

% Output
\draw [arrow] (fn-o) -- (out);
% ******************************** %
% ******************************** %
\end{tikzpicture}